\documentclass[10pt]{elsarticle}

 \usepackage{url}
 \usepackage{multirow}
 \usepackage{color}
 \usepackage{subfigure}
 \usepackage{algorithm}
 \usepackage{algorithmic}
\usepackage{verbatim}

\newcommand{\cps}{{\em critical points}}
\newcommand{\cp}{{\em critical point}}
\newcommand{\primitives}{{\em primitives}}
\newcommand{\primitive}{{\em primitive}}
\newcommand{\Figure}{{Fig.}}
\newcommand{\mynumber}{$100$}
\newcommand{\charaa}{{\em u}}
\newcommand{\charka}{{\em k}}
\newcommand{\eight}{{\em eight}}
\begin{document}

\begin{frontmatter}
%\newenvironment{comment}{Note:}
%\newenvironment{comment2}[1] {{Different Begin Comment:}} {{End: Comment}}
%\newcommand{\sunil}[2]{{\color{red} {\sc #1?:}} {\color{blue} {#2}}}

%\title{A Robust OHCR Framework for Devanagari On-line Handwriting Recognition} %\thanks{Grants or other notes
%\title {On-line Devanagari Handwritten Stroke Recognition Engine(based on Fuzzy Directional Features and Support Vector Machines)
%and a Robust OHCR Framework for Indian Languages}
\title{A Framework for On-Line Devanagari Handwritten Character Recognition}




\author{Sunil Kumar Kopparapu, Lajish VL }
\address{TCS Innovation Labs - Mumbai, \\ Tata Consultancy Services,
Pokhran Road 2, Thane (West),\\ Maharastra 400 601. INDIA.\\
SunilKumar.Kopparapu@TCS.Com}

%\maketitle

\begin{abstract}
The main challenge in on-line handwritten character recognition in Indian language 
is the large size of the character set, larger similarity between different
 characters in the script and the huge variation in writing style. 
In this paper we propose a framework for on-line handwitten script recognition
 taking cues from speech signal processing literature. The framework is based on
identifying strokes, which in turn lead to recognition of handwritten on-line characters rather that
the conventional character identification. Though the
framework is described for Devanagari script, the framework is general and can
be applied to any language.
 
The proposed platform consists of pre-processing,
feature extraction, recognition and post processing like the
conventional character recognition but applied to strokes.
The on-line Devanagari character recognition reduces to one of recognizing one
of $69$ \primitives\ and recognition 
of a character is performed by recognizing a sequence of such \primitives. 
We further show the impact of noise removal on on-line raw data which is usually
noisy. The use of Fuzzy Directional Features to enhance the accuracy of
stroke recognition is also described.
The recognition results are compared with commonly used directional features in literature 
using several classifiers.
\end{abstract}

\begin{keyword}
On-line handwriting recognition \sep Pre-processing \sep smoothing \sep knotless 
spline \sep curvature points \sep Fuzzy Directional Features \sep Dynamic Time
Warping \sep Support Vector Machine
\end{keyword}

\end{frontmatter}

%-----------------------Introduction--------------------------------------------

\section{Introduction}
\label{intro}

In recent times 
PDAs, palms and handheld PCs are more frequently \cite{joshi} 
being used for composing  short messages and e-mails. While composing messages
on these devices using conventional keypads is difficult because of their small form factor, 
these devices come
equipped with significant computing power making recognizing handwritten message
composition on board the device possible. While use of keyboard for composing English
language may be difficult but feasible, 
message composition through handwritten characters becomes the only
feasible way to compose messages in most Indian languages.
Compactness of devices are paving way to electronic pen (e-pen) and/or a stylus touching a 
pressure sensitive surface are some of the popular non-keyboard data entry devices that is
gaining popularity \cite{non_kboard_1,non_kboard_2,non_kboard_3}. 
 These input devices capture the pen lifts and the trace of the pen movement 
thus capturing handwritten strokes which is essentially a trace of the pen between a
pen down and a pen up. These traced strokes can then be converted into
electronically transferable character string using On-line Handwritten
Character Recognition (OHCR) algorithms.
Typically a trace of a pen between a pen-down and
a pen-up is a set of $x, y$ points which is 
 uniformly sampled in time (typically at $100$ Hz). These set of $x,y$ points
are  non-uniformly sampled in space.

\begin{figure}
\centering
\includegraphics[width=0.90\textwidth]{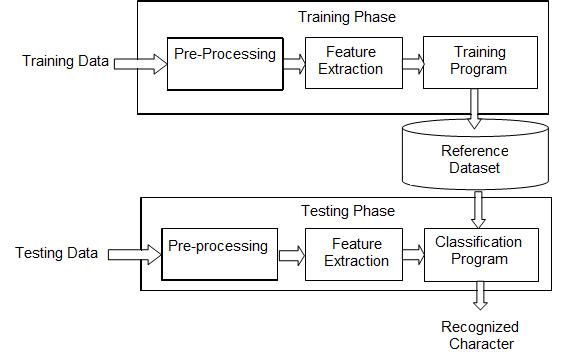}
\caption{High Level Block Diagram of a typical OHCR system}
\label{fig:hl_ohcr}
\end{figure}

An OHCR system (see \Figure\ \ref{fig:hl_ohcr})  generally consists of a learning phase and a testing
phase. In the learning phase the system learns and builds reference models for
all possible characters that need to be recognized. 
In the testing phase the character is compared
with the reference models of the character using a classifier to determine the
best matching reference character. 

The choice of the feature set to represent the on-line character determines
the ability of OHCR algorithm to distinguish one character from another while
being able to 
cluster together the same characters written
at different times and by different people.  
A typical OHCR recognition process would
consist of a pre-processing module followed by feature extraction and a
suitable classifier. 

On-line $x,y$ data acquired for the purpose of OHCR is most often uniformly sampled in 
time, making the captured $x,y$ data non-uniform in space. The outcome of this
is that the data points representing a character is dependent on the time
taken to write the character, making the number of data points representing a
character different for the same character and of the same size. To overcome
this non-uniform number of $x,y$ points representing a character many popular 
OHCR approaches make sure that the on-line data is pre-processed so as to obtain data that is uniformly sampled in space 
\cite{aparna,deepu_tamil,joshi}. While there are benefits in terms of the type of classification 
tools that one can use, it has a fundamental problem of not being able to
exploit the 
crucial curvature information that is embedded in the on-line data that 
is uniformly sampled in time. In this paper, we stick to the original 
uniformly sampled in time data produced by an e-Pen.
The difference between uniformly 
sampled in space and uniformly sampled in space is that in the later 
case every stroke or character has the same number of $(x, y)$ points, 
while when uniformly sampled in time the number of $(x, y)$ points are 
different.

\begin{figure}
\centering
\includegraphics[width=0.90\textwidth]{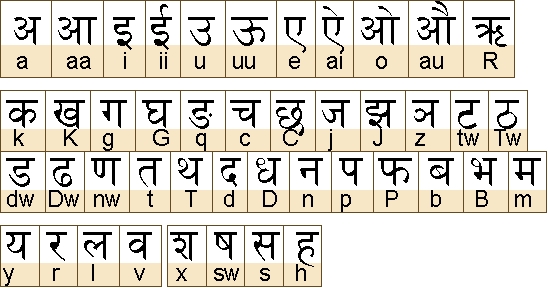}
\caption{Devanagari alphabet set \cite{web:alphabet}.}
\label{fig:alphabets}
\end{figure}

Devanagari script is a widely 
used Indian script being used by more than $500$ million people. It consists 
of %$14$ 
vowels and 
%$34$ 
consonants as shown in \Figure\ \ref{fig:alphabets}. It is used as the writing system for 
over $28$ languages including Sanskrit, Hindi, Kashmiri, Marathi and 
Nepali \cite{wiki:devanagari}. Devanagari is written from left to right in horizontal 
lines and the writing system is alphasyllabary. In English script 
barring a few alphabets, all the alphabets can be written in a single 
stroke\footnote{A stroke is defined as the resulting 
trace between a pen-down and its adjacent pen-up}. In contrast, in most Indian 
languages, characters are made up of two or more strokes. This writing
requirement makes it 
necessary to analyze a
 sequence of adjacent strokes to identify a character. 

\begin{figure}
\centering
\includegraphics[width=0.15\textwidth]{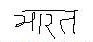}
\caption{A typical word has a horizontal line called {\em shirorekha}}
\label{fig:word}
\end{figure}

In Devanagari, like in most Indian languages,  for a 
consonant  vowel combination, the vowels are orthographically indicated 
by signs called {\em matras}. These modifier symbols are normally 
attached to the top, bottom, left or right of the base character which is
highly dependent on the consonant  vowel pair. In Indian languages,  
the consonants, the vowels, the matras and the consonant/vowel modifiers constitute 
the entire alphabet set. These composite characters are then joined 
together by a horizontal line, called {\em shirorekha} (see \Figure\ 
\ref{fig:word}) to form words.

\begin{comment}
\sunil{Consistency}{If 52 is "hk" then 4 should also be "hk"; if 53 is "K" then
2 should be "I" and 65 should be "Y". }
\end{comment}

\begin{figure}
\centering
\subfigure[Primitives (Characters)]{
\includegraphics[width=0.45\textwidth]{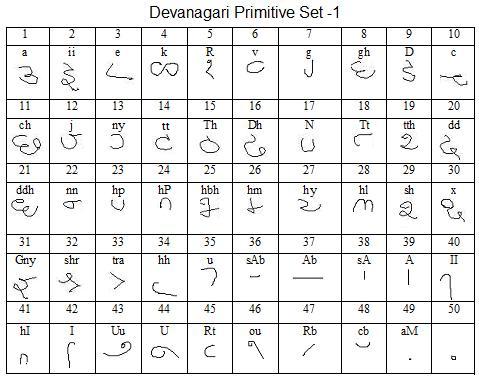}
\label{fig_char_prim}}
\subfigure[Additional Primitives]{
\includegraphics[width=0.45\textwidth]{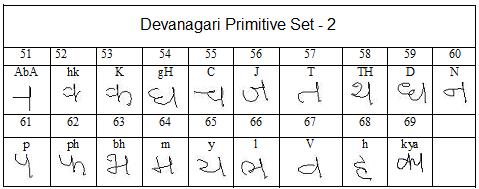}
\label{fig_sym}}
\caption{Devanagari \primitives\ (basic strokes)}
\label{fig_primitives}
\end{figure}

\begin{figure}
\centering
\includegraphics{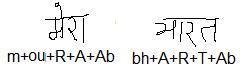}
\caption{Character formation using primitives}
\label{fig:eg}
\end{figure}

Devanagari character\footnote{we use character and alphabet interchangeably in
this paper} is made up of multiple strokes, so the identification of an
alphabet can be achieved by 
recognizing multiple strokes that make an alphabet. We earmarked, through visual 
inspection of handwritten Devanagari script, a basis like set of $69$ strokes
which we 
call them as \primitives. The set of all \primitives\ are shown in \Figure\ 
\ref{fig_primitives}.  The set of $50$ primitives (\Figure\ 
\ref{fig_char_prim}) are good enough to construct the entire 
character set in Devanagari. But we found that an additional set of $19$
\primitives\  
(\ref{fig_sym}) are often used by writers\footnote{a set of \mynumber\ hand
written paragraphs by different people were analyzed} in Devanagari script. These set of $69$ 
\primitives\ are sufficient to generate Devanagari script. Note that 
these \primitives\ can be combined together so as to form all the 
characters in Devanagari script. \Figure\ \ref{fig:eg} shows an example 
where \primitives\ (m, ou, R, A, Ab, $\cdots$ ) are combined together to 
form characters,  resulting into words. In this paper we assume that we 
can recognize the Devanagari characters by recognizing the \primitives\ and
analyzing a sequence of \primitives\ to identify a character.

In an unconstrained handwritten 
script these \primitives\ exhibit large variability
even for the same writer
making the task of recognizing \primitives\ and hence 
characters difficult. In this paper we 
use these \primitives\ as the units for recognition taking parallel from 
the recognition of {\em phone set} used in speech recognition literature.
\begin{comment}
\sunil{Query}{Is this challenge still there if we look at \primitives}
\end{comment}
The main contribution of this paper is 
\begin{enumerate}

\item Identification of a set of primitives which encompass Devanagari script,
\item Development of a framework to enable Devanagari script recognition by
recognizing primitives,
\item Use of pre-processing techniques to enhance primitive stroke recognition
accuracies,
\item Use of fuzzy directional feature (FDF) set to represent the on-line
characters \cite{pr_laj10} and
\item Use of relative spatial position of the primitives to enhance recognition.

\end{enumerate}

The rest of the paper is organized as follows. We review the state of the art 
OHCR for Indian
languages in Section \ref{sec:review} followed by the proposed OHCR framework 
in Section \ref{sys_arch}. 
The pre-processing techniques for noise removal of the on-line handwritten data 
are described in Section \ref{sec:pre-process} which also discusses the
procedure for identifying the critical points 
and gives an analysis of number of critical points identified on each
primitive.
We describe the feature extraction techniques including Directional Features
(DF), Extended Directional Features (EDF) and our \cite{pr_laj10} Fuzzy Directional Feature (FDF)
in Section \ref{f_extra}. 
The recognition experiments conducted for stroke level recognition using different classifiers including second order 
statistics based classifier, Discrete Time Warping (DTW) and Support Vector
Machine (SVM) are described in Section \ref{sec:experiments}.
We conclude in Section \ref{sec:conclude}.

%----------------------- OHCR review----------------------------------------------------------

\section{Indian language OHCR - An Overview}
\label{sec:review}

We give an overview of the recent advances, new trends
and important contributions in the area of OHCR for Indian languages. 
On-line handwriting recognition is of prime importance especially in the context of Indian
languages because of the fact that entering Indian language scripts is both
difficult and time consuming\footnote{A SIG on Indian Language SMS has been
working on various issues related to composing an SMS message in Indian
languages on a mobile device; however the concentration has been on using the
mobile keypad so far.}. Currently, word processing in Indian languages can be a vexing
experience, considering the restriction on use of the regular keyboard, designed for
English. Elaborate keyboard mapping systems are normally used in case of Indian
languages, which are not convenient to use.
While a large amount of OHCR literature is 
available for on-line handwriting recognition of English, Chinese and Japanese languages, 
relatively very less work has been reported for the recognition of 
Indian languages. In the case of Indian languages, research work has been active for Devanagari 
\cite{joshi,namboodiri}, Bangla\cite{parui,bhatt}, Tamil
\cite{niels,aparna,joshi_tamil,sundar,toselli} and Telugu
\cite{babu,pvs}.

Joshi et al \cite{joshi} describe a system for the automatic recognition of isolated
handwritten Devanagari characters obtained by linearizing consonant conjuncts.
They used structural recognition techniques to reduce some characters to others. 
The residual characters are then classified using the subspace method. Finally the
results of structural recognition and feature based matching are mapped to give
a final output 
and the system is evaluated for writer dependent scenario.
In another work \cite{ranade}, a set of stroke templates is derived from analysis of 
common writing styles 
of different Devanagari characters, and each character 
represented by a set of combinations of these stroke
templates. In \cite{connell} the authors use a combination of two HMM classifiers 
trained with on-line 
features and three Nearest Neighbor classifiers each trained on different sets of
on-line features for Devanagari character recognition. The combination of
on-line
and off-line classifiers is shown to improve the accuracy from 69.2\% (on-line, HMM
alone) to 86.5\%. 

Tamil on-line handwriting
recognition has also been attempted with varying degree of success. In \cite{keerthi} the problem
of representation of Tamil characters is considered. In another work \cite{deepu} the
subspace based on-line recognition of Tamil and other Indian languages is described.
Joshi et al \cite{joshi_tamil} use template based elastic matching algorithms
for  on-line Tamil handwriting recognition 
%based on template based elastic matching algorithms. 
They use the advantage
of elastic matching algorithms which do not require a very large amount 
of training data, making them suitable for writer dependent recognition.
Aparna et al \cite{aparna}, represent Tamil character strokes 
as strings of shape features. 
In order to recognize an unknown stroke, its equivalent feature string is
computed. The test stroke is then identified by searching the database using a
flexible string matching algorithm. Once all the strokes in the input are known,
the character is determined using a Finite State Automaton.
Prior knowledge about popular writing styles has been exploited to design a first stage 
classifier for Tamil characters in  \cite{sundaram}. The authors observe that
the start of any Tamil character is either a line, semi-loop or a loop. 
Accordingly, the candidate
choices are pruned during recognition. 
Artificial Neural Network (ANN) based approach is also proposed 
\cite{sundar} for the recognition of on-line Tamil characters.
In another effort on Tamil character recognition \cite{niels}, templates are identified
from the training set using Agglomerative Hierarchical Clustering and Learning
Vector Quantization (LVQ) with dynamic time warping (DTW) as the distance measure. A DTW-based
Nearest Neighbor classifier is then employed for matching the test sample. 

In \cite{bhatt_bangla}, the authors describe a novel direction code based feature 
extraction approach 
for recognition of on-line Bangla handwritten basic characters. 
It is a $50$-class recognition problem and 
they achieved $93.90$ \% and $83.61$ \% recognition accuracies on  training and test sets respectively.
For the problem of Telugu character recognition, PVS Rao et al \cite{pvs} perform 
coarse matching
with the templates using the number of $x-y$ extrema points in the test sample
and affine matching using DTW. HMMs have also been used for Telugu \cite{babu} on-line 
handwritten character recognition.
In \cite{kunte}, feed-forward Neural Networks with a single hidden layer have
been used for the 
recognition of handwritten Kannada characters. The authors use approximation coefficients derived 
from Wavelet decomposition on the pre-processed $(x,y)$ as 
features for representing the characters.

\section{Proposed System Architecture}
\label{sys_arch}

\begin{figure}
\centering
\includegraphics[width=0.90\textwidth]{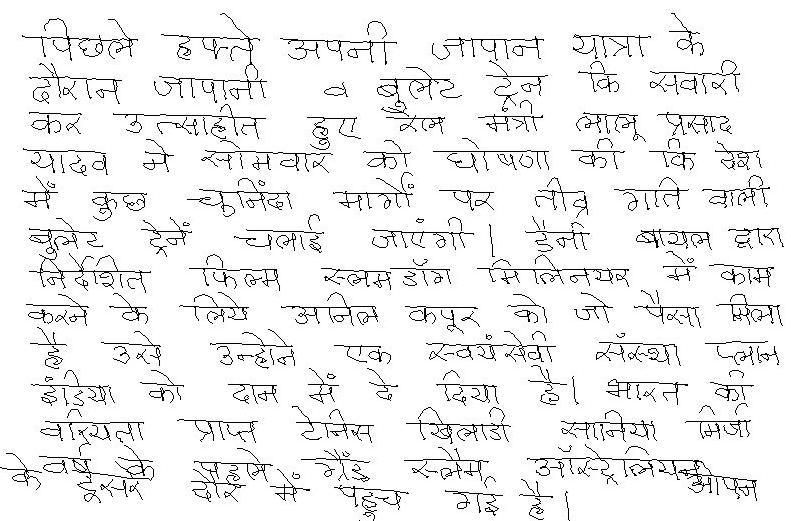}
\caption{Sample on-line handwritten paragraph data }
\label{fig_para}
\end{figure}

\begin{figure}
\centering
\centerline{\includegraphics[width=0.90\textwidth]{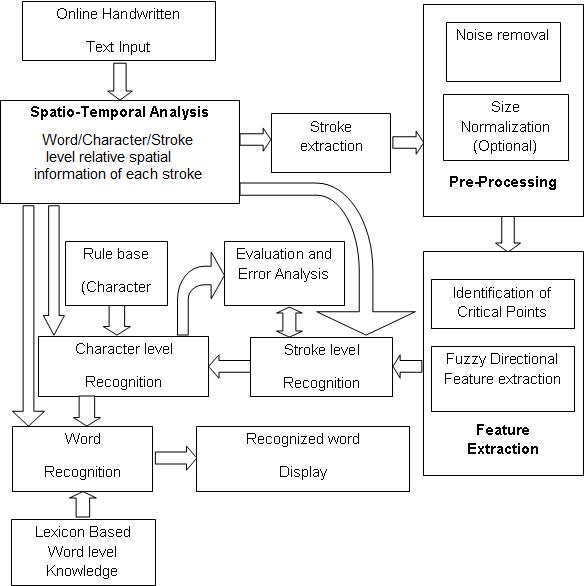}}
\caption{Proposed System Architecture}
\label{fig:sys_arch}
\end{figure}

We proposed a frame work for Indian language OHCR system (\Figure\
\ref{fig:sys_arch})
where we can use the stroke 
recognition for Devanagari handwritten script recognition. 
The stroke recognition would be language independent, though the shape of the
strokes and the number of strokes might vary from one language script to
another language script.
Initially
the on-line data (see \Figure\ \ref{fig_para}) is acquired and 
a spatio-temporal analysis of the individual strokes is done. Typically, this
analysis\footnote{based on the understanding of the written script} provides the ability to segment the paragraphs into words (based on
{\em shirorekha} identification); identify {\em matras} by identifying the
relative position of the strokes. This analysis can be used to improve the
performance of the stroke recognition. For example, having identified a {\em
matra} based on the spatial position of the stroke, we could constrain the
recognition to  only the reference {\em matras}. 
The individual strokes are  then recognized to be one of the $69$ primitives. 
The actual stroke recognition has several steps; the stroke is first 
pre-processing, and features extracted from each stroke before stroke recognition.
The stroke level recognition is modified based on the failure of character
recognition to put together a sequence of strokes to form a character. The  
error analysis block helps in improving the stroke recognition.
Rules for character formation from 
a sequence of strokes and the spatio-temporal knowledge help in ascertaining if
the recognized sequence of strokes are valid or not.
The framework has a  word recognition  module which is 
very general and based on a lexicon making it adaptable for any Indian
language.
Lexicons and other language models are 
an important aspect of achieving acceptable accuracy for OHCR. Barring a few reported methods that use
lexicons, the use of language models has not been exploited substantially for
Indian scripts.
Word recognition uses the characters identified along with the 
spatio-temporal information and domain based lexicon word level knowledge.

The described framework in \Figure\ \ref{fig:sys_arch} is influenced by the
speech recognition literature. The strokes are analogous to phonemes in
speech. It is well know in speech literature that the phoneme recognition
accuracies are poor, however the final output of the speech recognition is
significantly high. The poor phoneme recognition in speech recognition is enhanced by lexicons and
language models. In a similar fashion, in the proposed framework we argue that the
use of lexicon will improve the OHCR. In this paper, we restrict our discuss to stroke
recognition.

%\section{Pre-processing}
\section{Stroke Recognition}
\label{sec:pre-process}
We analyzed handwritten script by \mynumber\ people and identified a set of
$69$ strokes (\Figure\ \ref{fig_primitives}) which we will refer to as \primitives) which encompassed the Devanagari script.
These  \primitives\ are shown in Fig \ref{fig:eg}. The rest of the paper deals
with the recognition of Devanagari script by the recognition of 
these  \primitives.

Let a  \primitive\ be represented by a variable number of 2D points which are
in a time sequence.
For example an on-line script would be represented as 
\begin{equation} \left \{(x_{t_1} ,
y_{t_1} ), (x_{t_2} , y_{t_2} ), \cdots , (x_{t_n} ,
y_{t_n}) \right \} \end{equation} $t$ denotes the time and assume that $t_1 < t_2 <
\cdots  
< t_n$ and $n$ represents the total number of points. Equivalently we can
represent the on-line data %(see Fig \ref{fig_sample}(a)) 
as
\begin{equation} \left \{(x_1 , y_1 ), (x_2 , y_2 ), \cdots , (x_n , y_n )
\right \} \end{equation} by
dropping the variable $t$.
The number of points denoted by $n$ vary depending on the size of the
\primitive\
and also the time taken to write the \primitive.
Most script digitizing devices (popularly called electronic pen) sample the
script uniformly in time. For this
reason, the number of sampling points is large when the writing speed is slow
which is especially true at curvatures.

\begin{figure}
\centering
\subfigure[Devanagari \charaa]{
\includegraphics[width=0.45\textwidth]{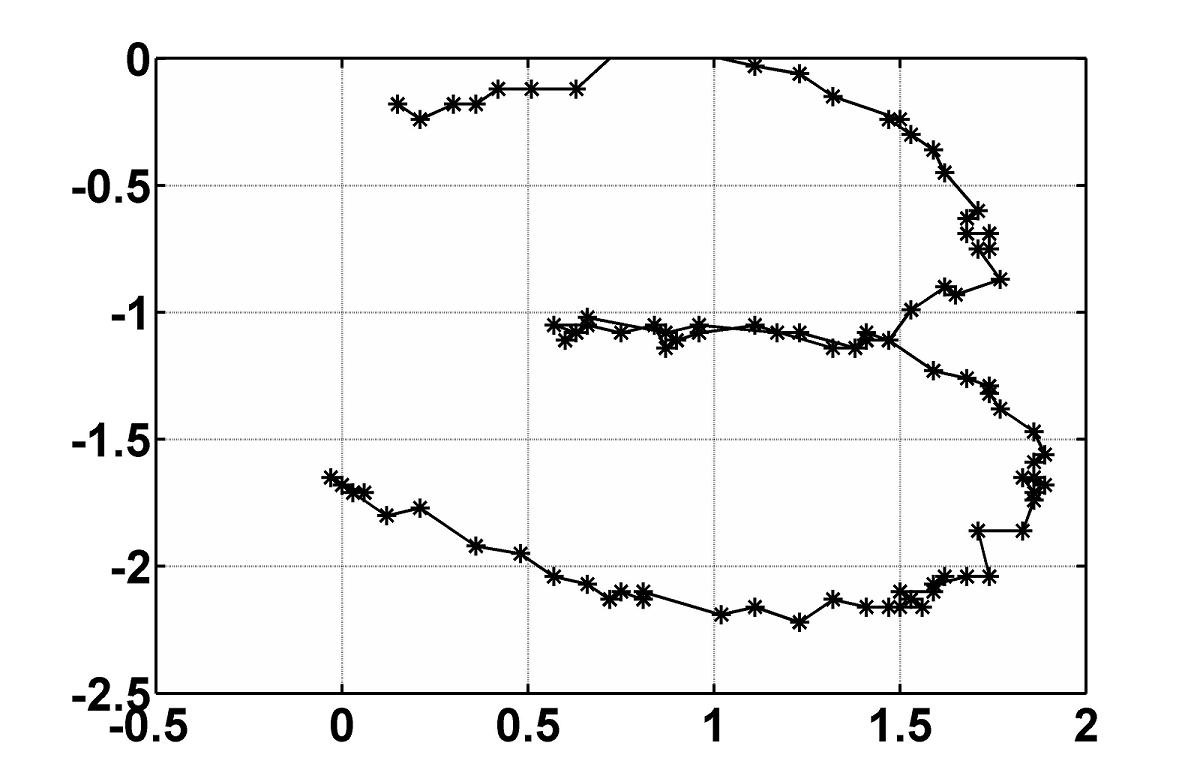}}
\subfigure[Devanagari \charka]{
\includegraphics[width=0.45\textwidth]{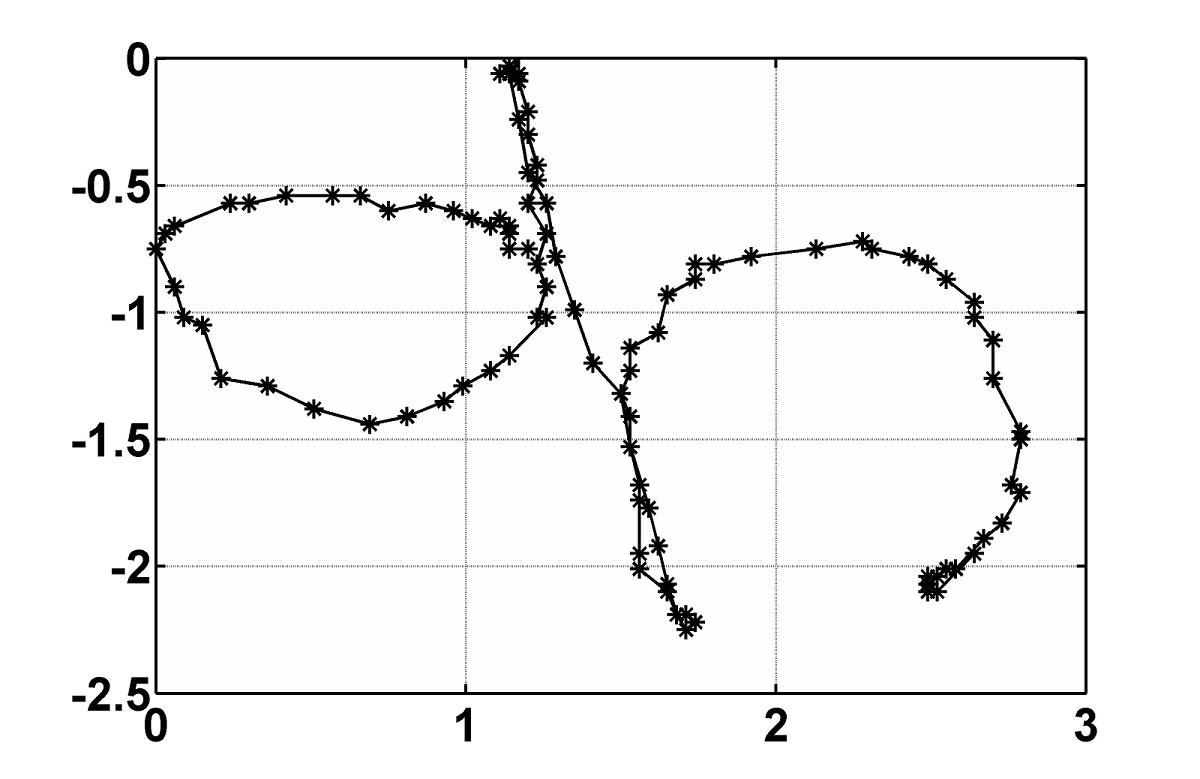}
}
\caption{Sample on-line Devanagari characters \charaa\ and \charka: Raw data }
\label{Fig:original}
\end{figure}
Noise in on-line script is inherent. As seen in \Figure\ \ref{Fig:original}
the handwritten character is far from smooth. 
The number of data points is very sparse especially when the pen movement is fast  
and when the pen movement is slow they are prone to be contaminated by high frequency noise.

There are
essentially two types of noise that contributes to the noisy data, (a)
the inherent shake of the hand of the writer especially at the beginning and end of the 
stroke and (b) contribution by the noise creeping in due to 
digitization process.
\Figure\ \ref{Fig:original} (a) and (b) show samples on-line Devanagari
primitives \charaa\ and \charka\ collected 
from a writer using Mobile e-Note Taker \cite{non_kboard_1}. We can observe that the characters are not smooth and
contaminated by noise. The noise severely affects the performance of on-line character recognition
algorithms. There are essentially two ways of taking care of the noise, (a) in the first case an appropriate noise
removal algorithm is used on the raw noisy data and (b) in the second case one uses a feature extraction algorithm
that can compensate for the noise.

Pre-processing is a necessary first step in OHCR. In the next sections we
discuss noise removal pre-processing techniques and show that noise removal
helps in improving stroke recognition.

\subsection{Smoothing based Noise Removal}
\label{sec:noise_rem}

Smoothing and filtering are the two important techniques used for
noise reduction. Smoothing usually averages a data point with respect to
its neighboring data points such that there is no {\em large} variation
between adjacent data points (for example \cite{smu_1,smu_3}). Filtering on the other hand
eliminates duplicate data points and reduces the number of points \cite{smu_6}. Some filtering techniques
force a minimum distance between adjacent data points
(for example \cite{smu_3,filt_2}) which tend to produce data
points that tends to be equally spaced. According to the writing speed,
the distance between the points may vary significantly and interpolation
can be used to obtain uniformly spaced  points. In some filtering,
minimum change in the direction of the tangent to the drawing for
consecutive points are maintained \cite{filt_3}. This produces more
points in the greater curvature regions. Filtering can also be done by
the use of convolution with one dimensional Gaussian kernels
\cite{filt_4}, which reduces the noise due to pen movement and errors
in the sensing mechanism. Joshi et al. \cite{joshi} reduced the effect
of noise with the help of 5-tap low pass Gaussian filter, where each
stroke is filtered separately. Malik et al. \cite{filt_5} proposed time
domain filter by using the convolution of input sequence with a finite
impulse response for smoothing a jitter appeared in the sequence.
Although many techniques exist in literature to suppress noise,
it is very difficult to select a technique such
that it can work equally for all types of strokes.
In some other studies smoothing and filtering are performed as part of single operation. An
example of this is piecewise-linear curve fitting
\cite{filt_6,filt_7}.

\subsubsection{Wavelet based noise removal}
\label{wavelet_nr}

Here as part of noise removal we performed smoothing on the raw data using 
Discrete Wavelet Transform (DWT) based decomposition 
using Daubechies wavelet\footnote{We do not dwell on this since this is well covered in Digital Signal Processing literature}.
The DWT decomposition helps in removing noise due to small undulation caused due to the sensitiveness of the sensors on the electronic pen 
and inherent shake while writing. 
\Figure\ \ref{Fig:dwt} (a) and (b) show the noise removal due to DWT
on the Devanagari
characters \charaa\ and \charka.

\subsubsection{Knotless Spline for noise removal}
\label{knotless_spline_nr}

This noise removal technique is based on splines.
Different spline (\cite{bk_sup_smu_1,bk_sup_smu_2}) based smoothening have been successfully applied for denoising
noise contaminated signal for example, \cite{anoop,n2}.
However, the degree of smoothness depends on the number and position of the
control points and chosen knots. If the knots are {\em close}
to each other, the smooth curve between the two knots would be linear. If the
knots are {\em far} apart, a higher order polynomial would be needed 
for fitting a smooth curve between the two knots. The knotless spline technique 
uses a cubic spline based polynomial 
approximation with the knots being selected automatically\footnote{both the number
of knots and the location of the knots}. Hence, the smoothing technique becomes a cubic polynomial
curve fitting with a variable span\footnote{span is defined as the distance
between the two consecutive knots} .

Let the sequence $(x^r_i, y^r_i)_{i=0}^{n}$
represents a handwritten stroke made up on $n$ points.
For the purpose of noise removal we treat the sequence $x^r_i$ and $y^r_i$
separately and remove noise from each of these sequences independently.
The noise removal process is described below for the sequence $x^r_i$.

\begin{figure} 
\centering
\subfigure[]{
\includegraphics[width=0.45\textwidth]{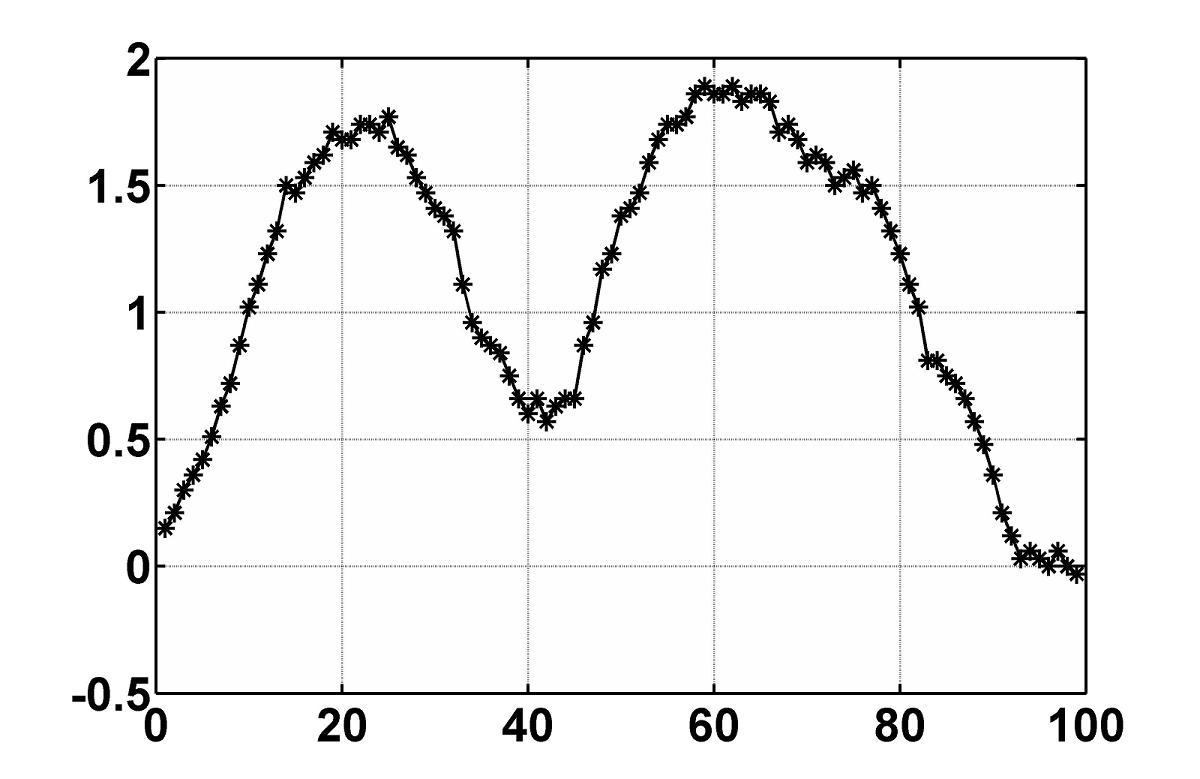}}
\subfigure[]{
\includegraphics[width=0.45\textwidth]{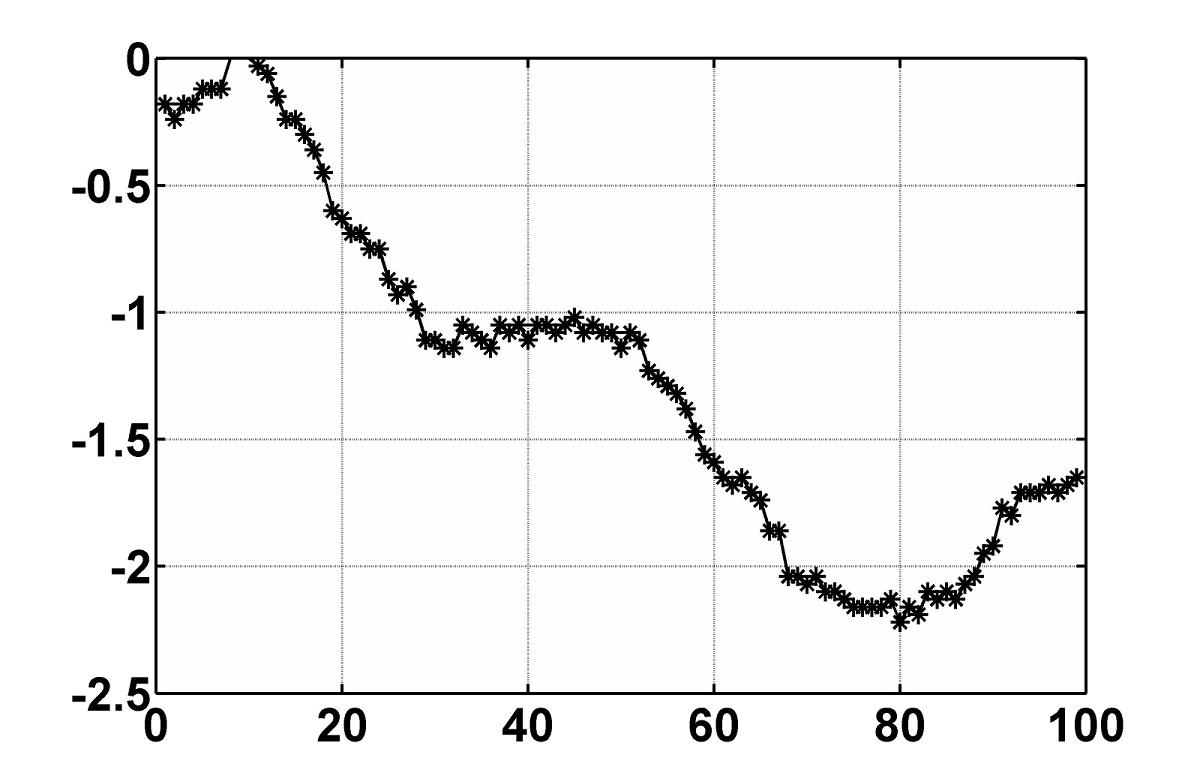}}
\caption{(a)-(b) $x$ and $y$ sequences of Devanagari characters \charaa: Raw data }
\label{Fig:compare_1}
\end{figure}

\begin{figure}
\centering
\subfigure[]{
\includegraphics[width=0.45\textwidth]{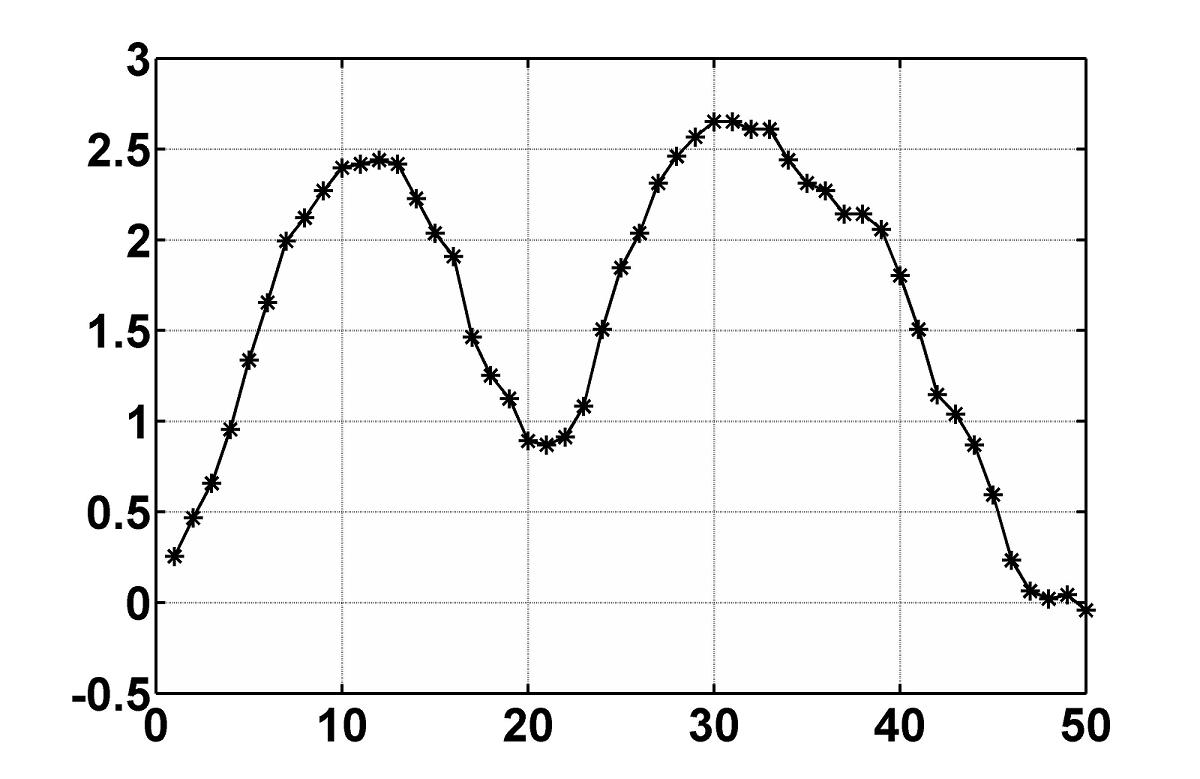}}
\subfigure[]{
\includegraphics[width=0.45\textwidth]{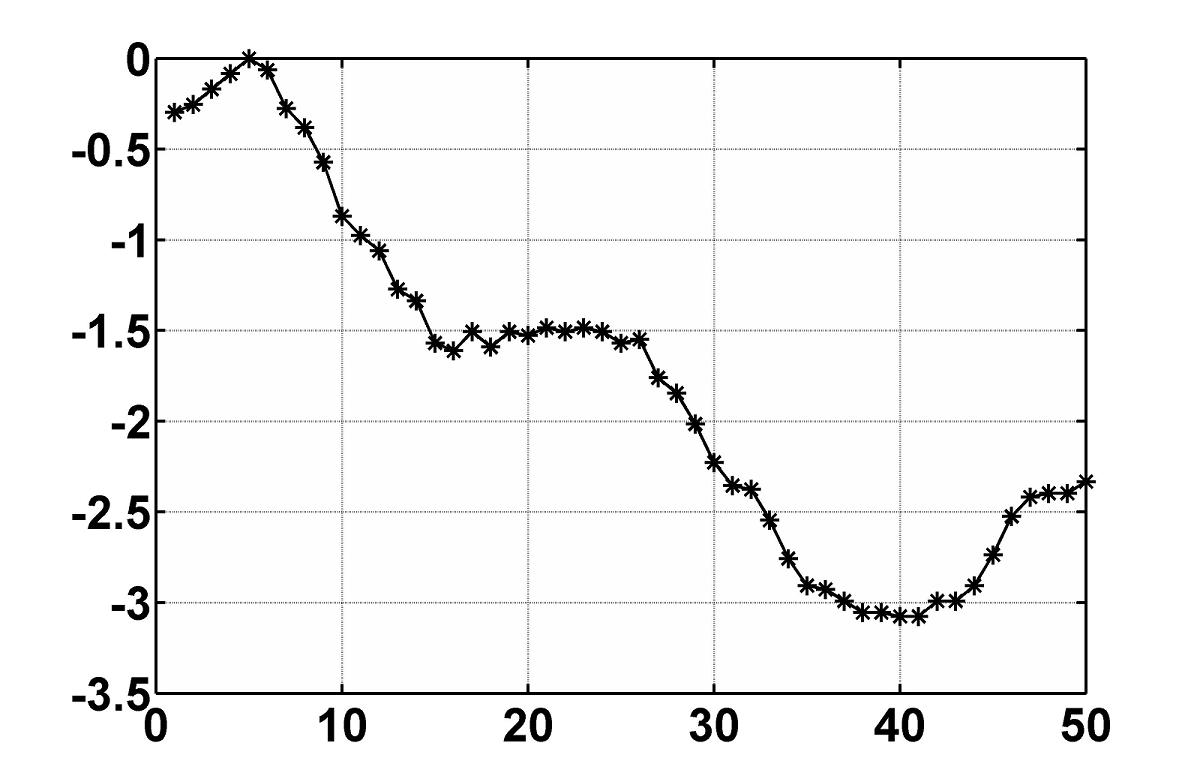}}
\caption{(a)-(b) $x$ and $y$ sequences of Devanagari characters \charaa : Smoothed
using DWT }
\label{Fig:compare_2}
\end{figure}

\begin{figure}
\centering
\subfigure[]{
\includegraphics[width=0.45\textwidth]{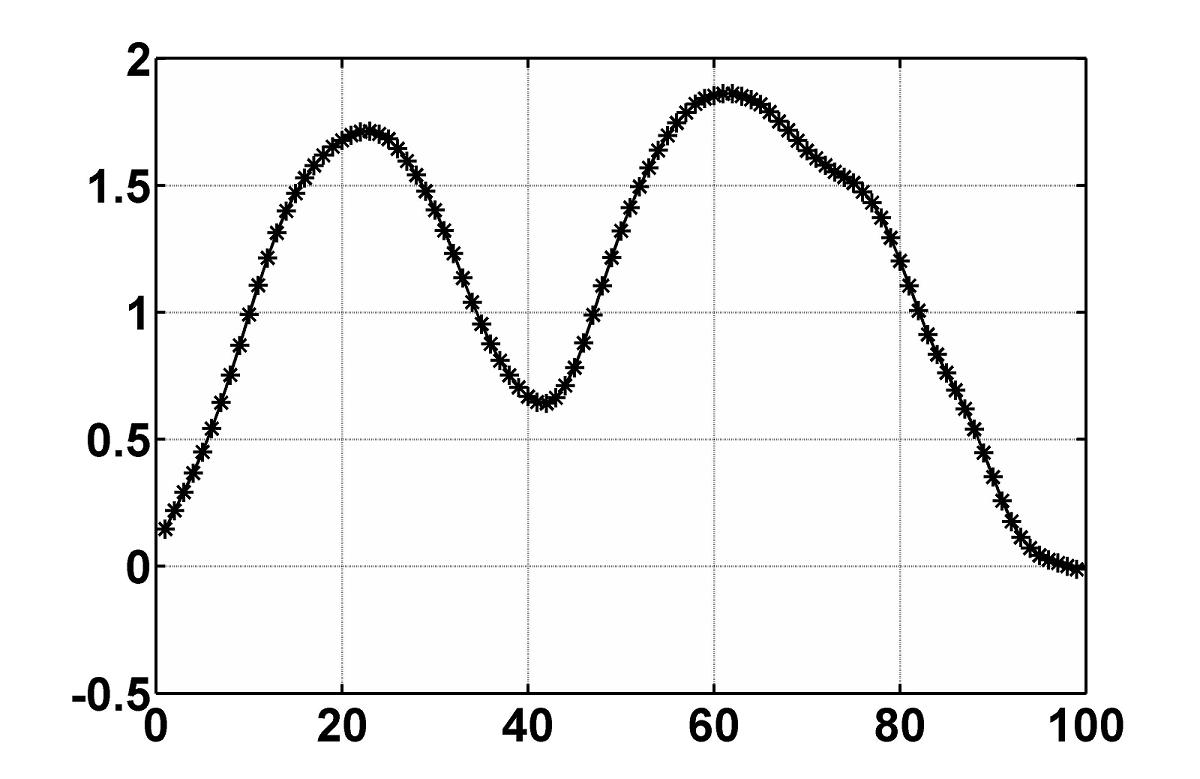}}
\subfigure[]{
\includegraphics[width=0.45\textwidth]{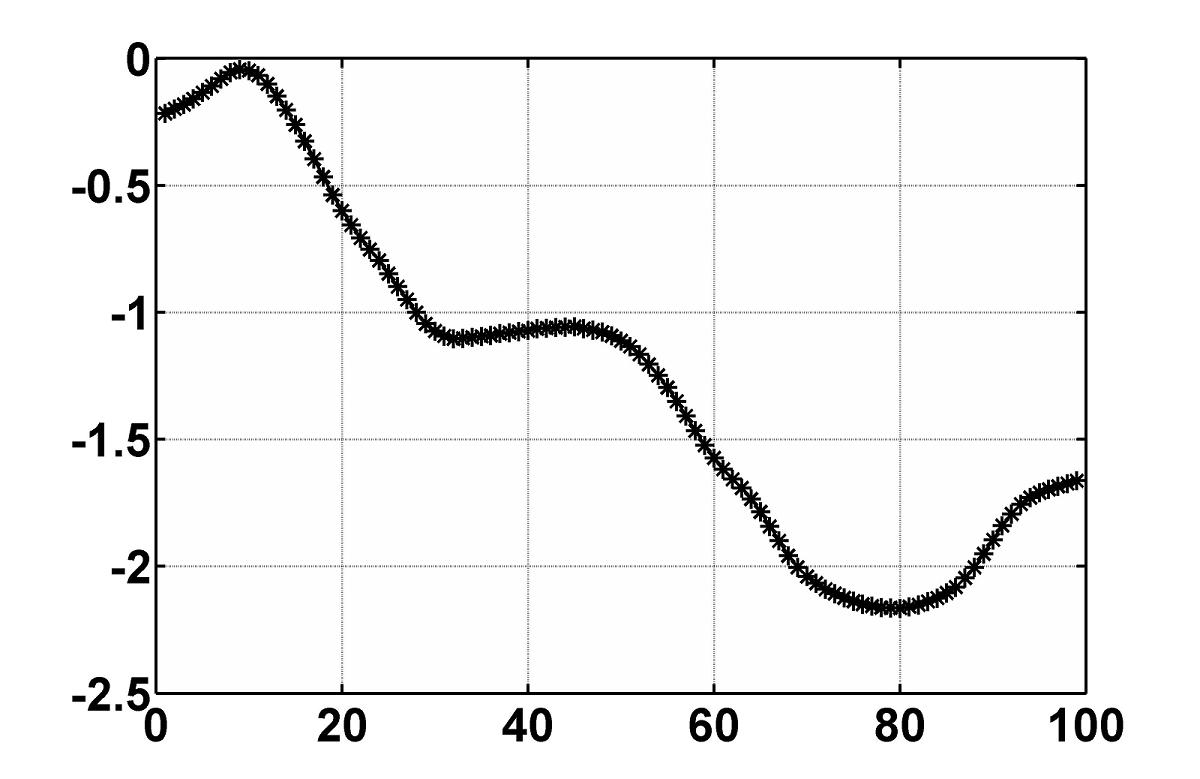}}
\caption{(a)-(b) $x$ and $y$ sequences of Devanagari characters \charaa : Smoothed
using the proposed procedure }
\label{Fig:compare_3}
\end{figure}

\begin{enumerate}
\item Set the span to be $n/2$ data points (consider only $n/2$ of the
original $n$ points, namely, $\{x^r_i\}_{i=0}^{n/2}$)
\item Fit a cubic spline in the span
%\footnote{span is defined as the distance between the two consecutive knots} 
compute and a mean squared error (MSE) is 
calculated between the fitted spline and the actual  data points, namely, find $\{a_i\}_{i=1}^{3}$ such that
$f(x^r_i) = a_0 + a_1 x_i^r+ a_2 {x^r}^2_i+ a_3 {x^r}^3_i$  such that
\begin{equation} MSE =
\sum_{i=1}^{n/2} (f(x^r_i)
- x^r_i)^2 \end{equation} is minimum
\item Reduce the span by 25\% (namely, consider  $\{x^r_i\}_{i=0}^{n/2-n/8}$)
and repeat Step 1 and 2 until the span is  20\% of the initial span. 
\item The span with smallest MSE is selected as the optimum span with the
starting and end points of the span are the chosen knots and a cubic spline is fitted in this span. 
\item Repeat on the remaining data points until all points are covered.     
\end{enumerate}
It is to be noted that this process automatically selects the number and the location of
the knots unlike other spline denoising techniques which requires the user to
specify the number of knots.
We compare the effect of noise removal using (a) wavelet denoising technique
and (b) the method proposed above; the results of smoothing the $x$  and $y$ sequence separately for 
the Devanagari character \charaa\ are shown visually in \Figure\  \ref{Fig:compare_1}, \ref{Fig:compare_2}
and \ref{Fig:compare_3}. 

\begin{figure}
\centering
\subfigure[]{
\includegraphics[width=0.45\textwidth]{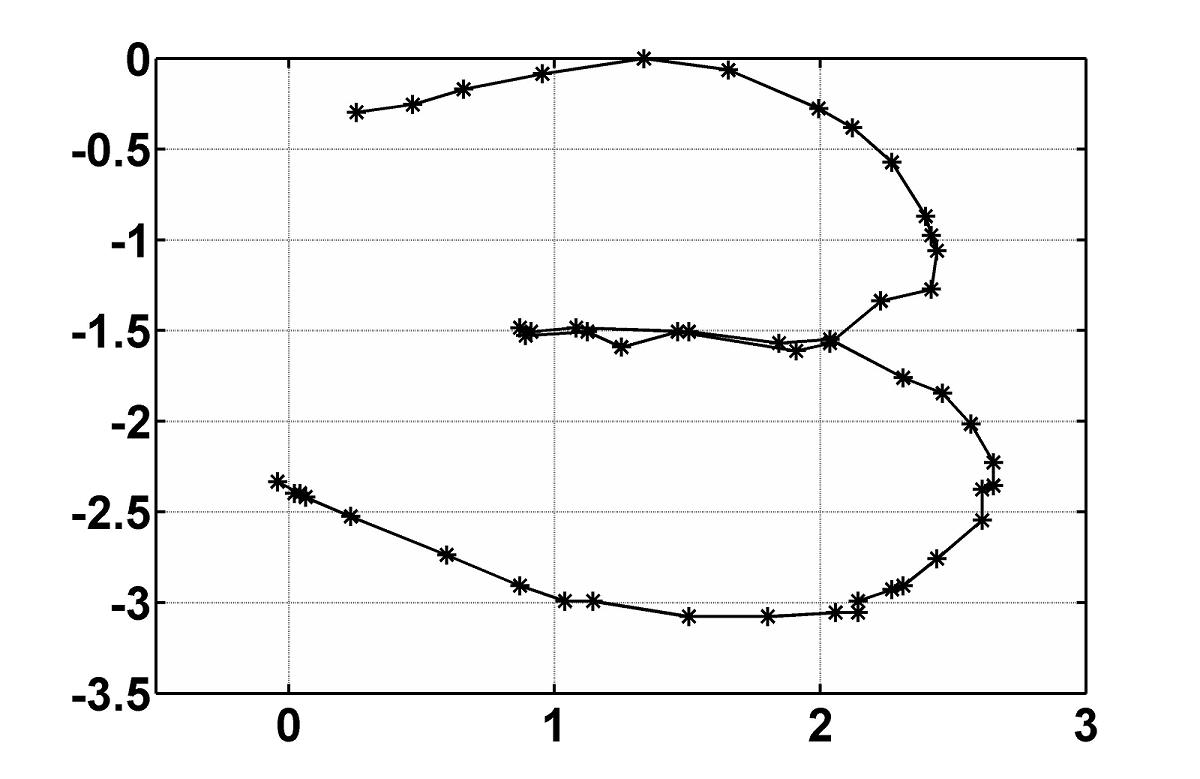}
}
\subfigure[]{
\includegraphics[width=0.45\textwidth]{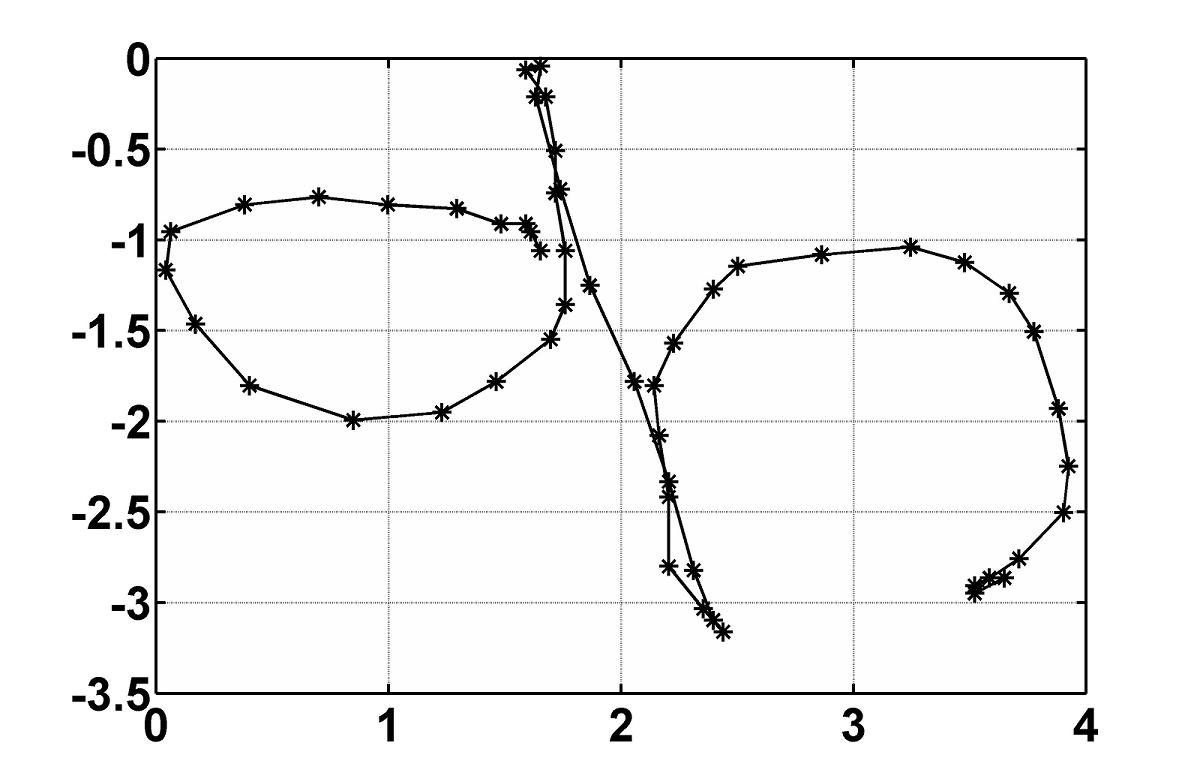}}
\caption{(a)-(b) On-line Devanagari characters \charaa\ and \charka:
Filtered using DWT }
\label{Fig:dwt}
\end{figure}
\begin{figure}
\centering
\subfigure[]{
\includegraphics[width=0.45\textwidth]{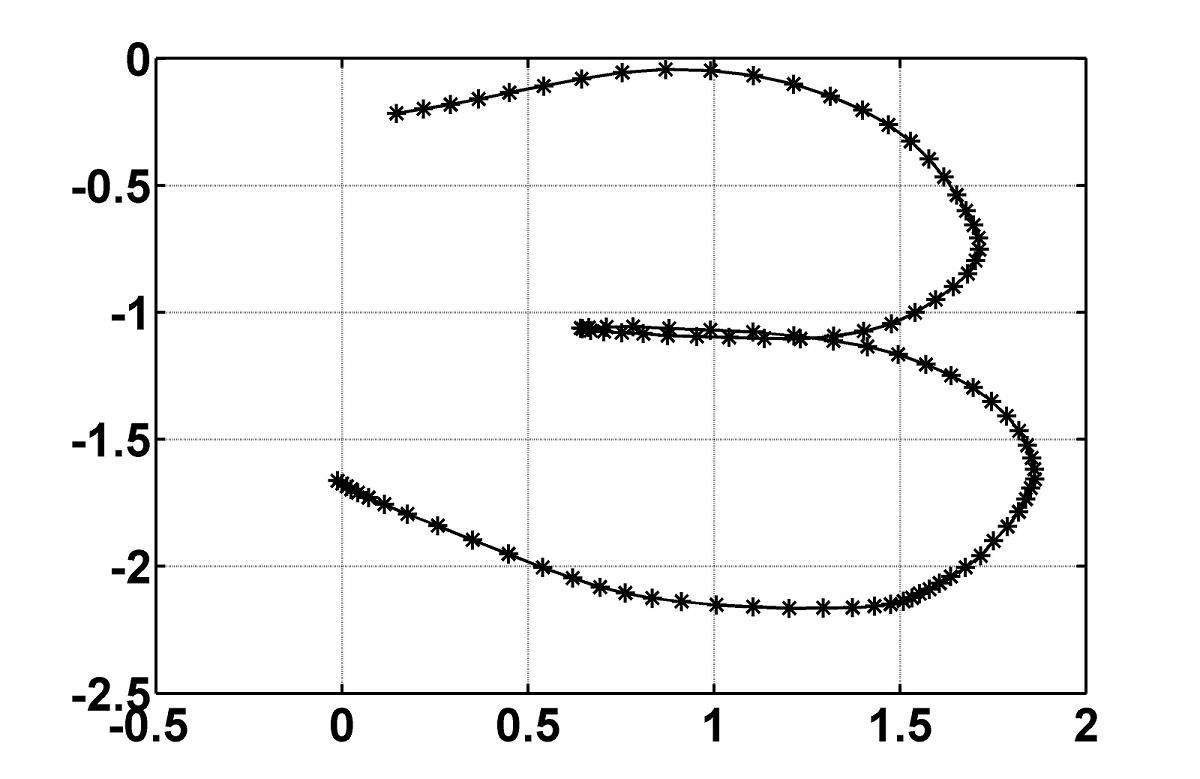}}
\subfigure[]{
\includegraphics[width=0.45\textwidth]{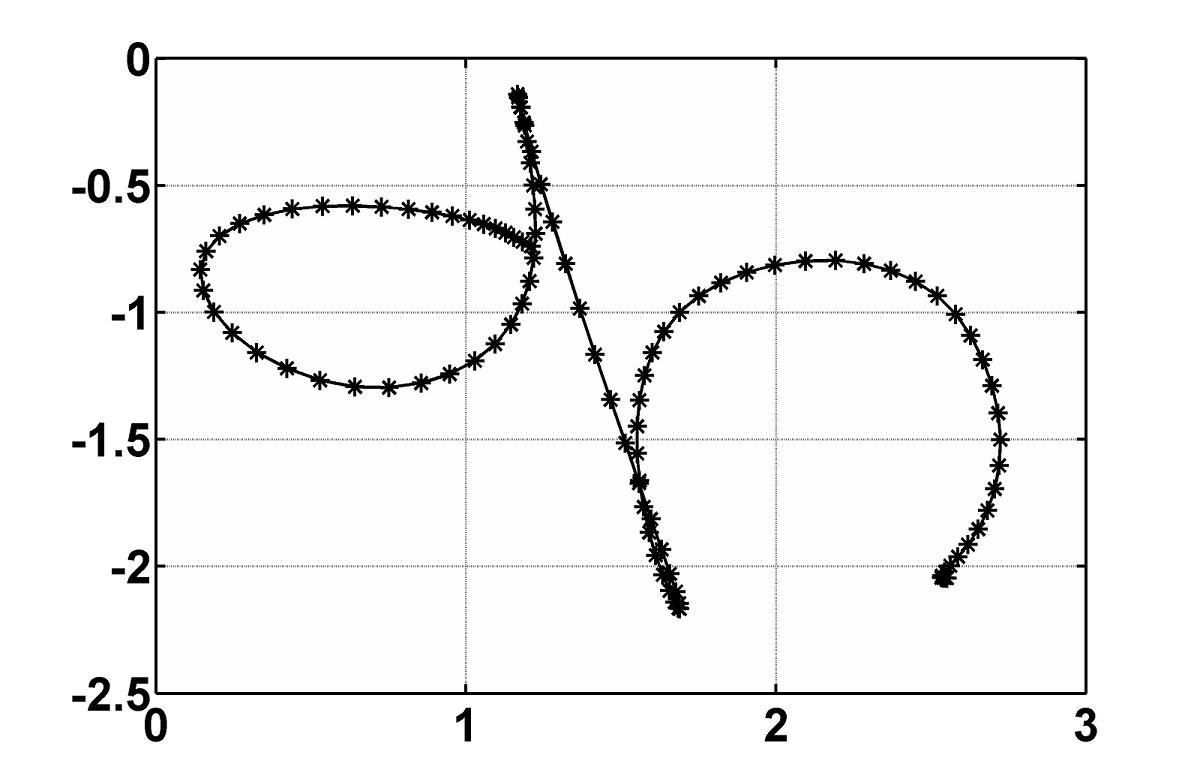}}
\caption{(a)-(b) On-line Devanagari characters \charaa\ and \charka: Smoothed
using proposed method}
\label{Fig:supsmu}
\end{figure}

\Figure\ \ref{Fig:dwt} and \Figure\ \ref{Fig:supsmu} shows on-line Devanagari characters
\charaa\ and \charka\ smoothed using DWT and knotless spline technique. 
It can be clearly seen that the noise removal is better
using the knotless spline denoising process compared to the wavelet based denoising.

We conducted a separate recognition experiment for evaluating the noise removal technique based on
(a) Raw data (without pre-processing) (b) DWT based denoised data
and (c) the knotless spline based pattern smoothing technique. The average
stroke recognition
accuracies obtained for $69$ Devanagari characters were about 51.21\% for raw data, 60.86\% for DWT based smoothed data and 
about 70.29\% for the data pre-processed using the knotless spline method. This
improvement in recognition
clearly demonstrates that noise removal is a necessary for improved performance
in OHCR.

%-------------------------Feature Extraction--------------------

\section{Direction Based Feature Extraction}
\label{f_extra}

Several temporal features have been used for script recognition in general 
and for on-line Devanagari script recognition in particular \cite{menier,garcia,schenk,conne}.
We discuss feature set based on directional properties of the curve 
connecting two consecutive critical points identified on a Devanagari
\primitive. In particular we discuss three different types of features based on
directional properties.

\subsection{Critical Point Extraction}
\label{sec:cr_point}

The curvature points (also called \cp) are extracted from
the smoothed handwritten data. The denoised sequence $(x_i, y_i)_{i=0}^{n}$
represents a noiseless handwritten stroke. We treat the sequence $x_i$ and
$y_i$ separately and calculate the critical points for each of these sequence. 
For the $x$ sequence, we calculate the first difference ($x'_i$) 
as \begin{equation} x'_i = sgn(x_i -
x_{i+1})\end{equation} where
\begin{eqnarray}
sgn(k) = +1 &\mbox{if} & x_i - x_{i+1} >0\\ \nonumber
sgn(k) = -1 & \mbox{if} & x_i - x_{i+1} <0 \\ \nonumber
sgn(k) = 0 & \mbox{if} &x_i - x_{i+1} =0 
\end{eqnarray}

\begin{figure}
\centering
\subfigure[]{
\includegraphics[width=0.45\textwidth]{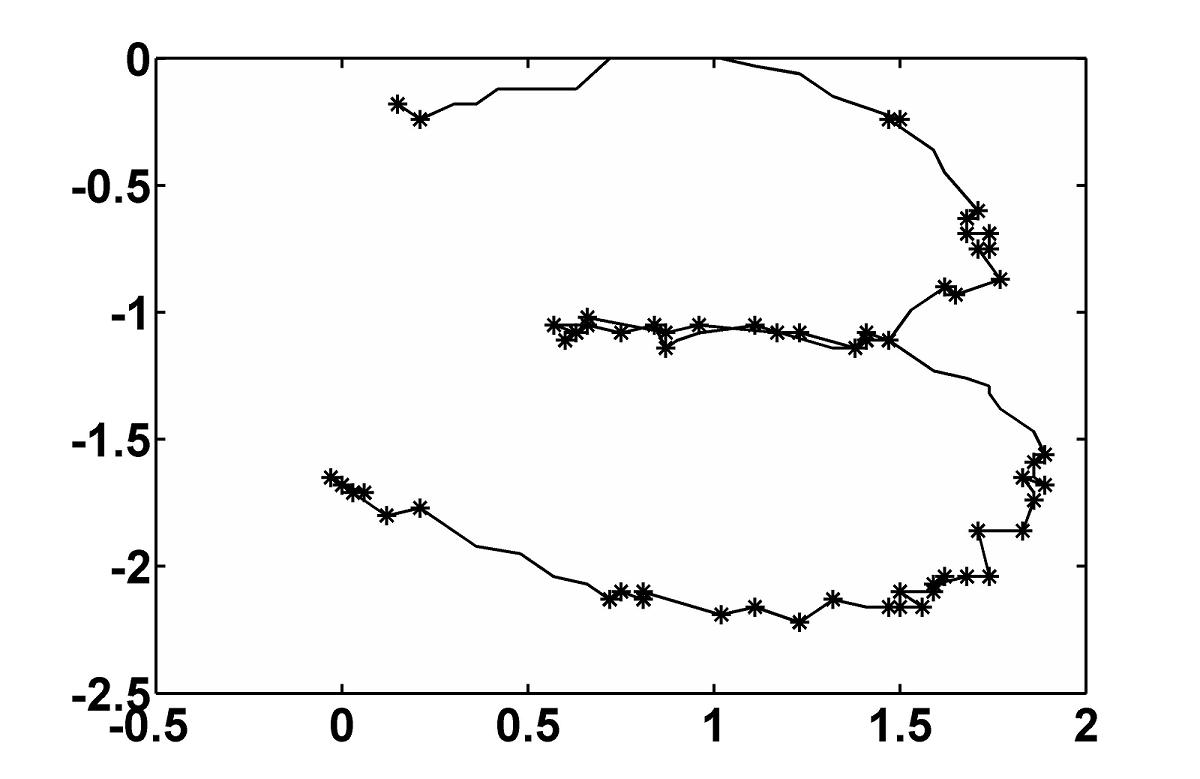}}
\subfigure[]{
\includegraphics[width=0.45\textwidth]{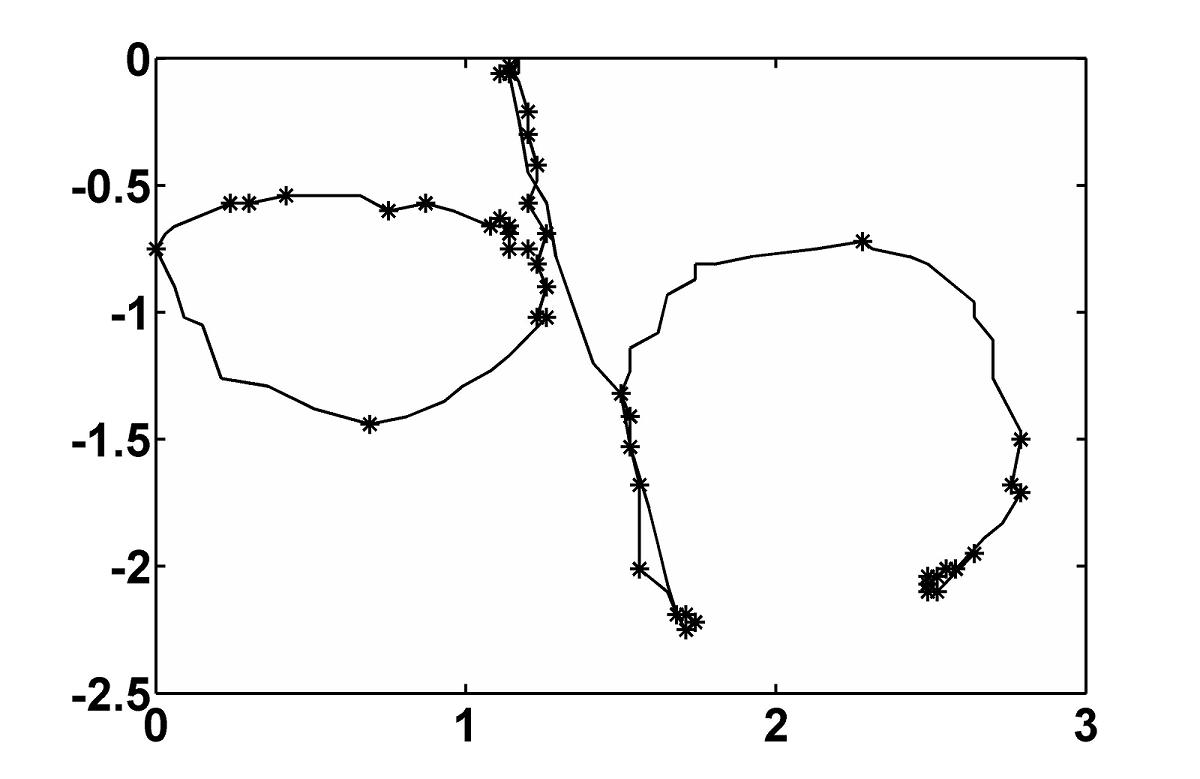} }
\caption{(a)-(b) Curvature points on on-line Devanagari characters \charaa\ and
\charka : Raw data}
\label{Fig:crt_pnt_org}
\end{figure}

\begin{figure}
\centering
\subfigure[]{
\includegraphics[width=0.45\textwidth]{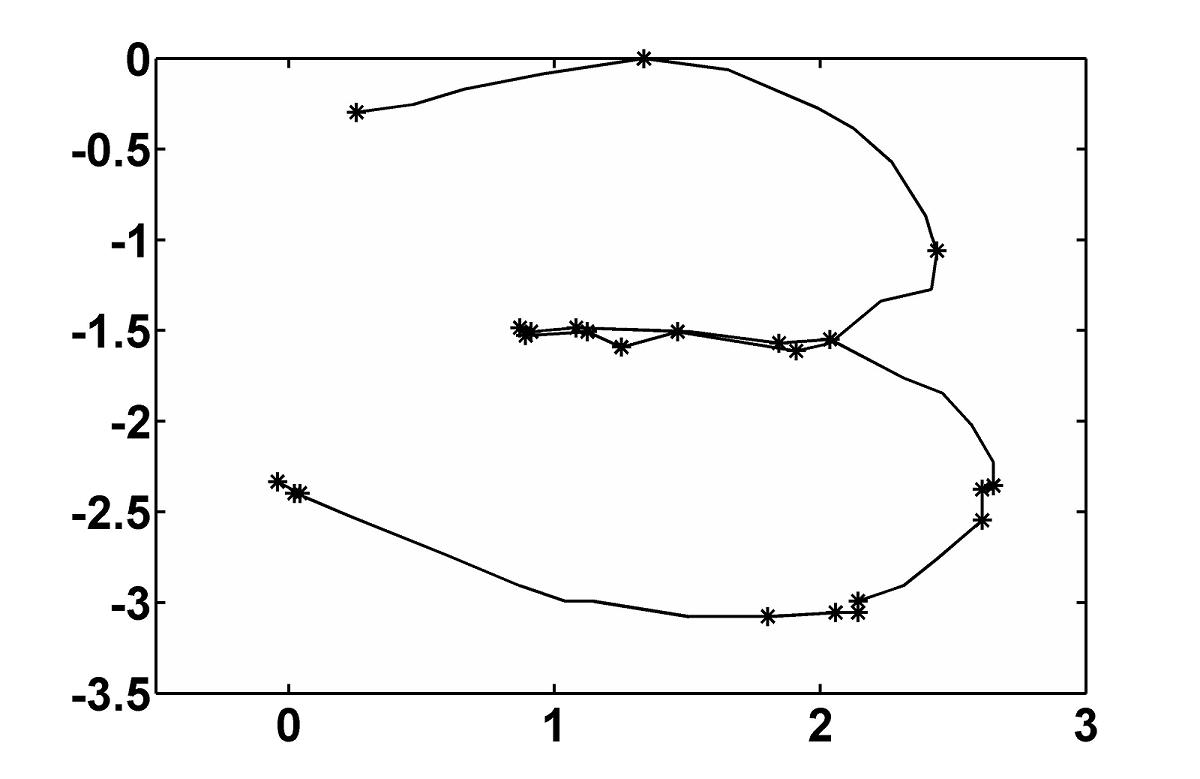}}
\subfigure[]{
\includegraphics[width=0.45\textwidth]{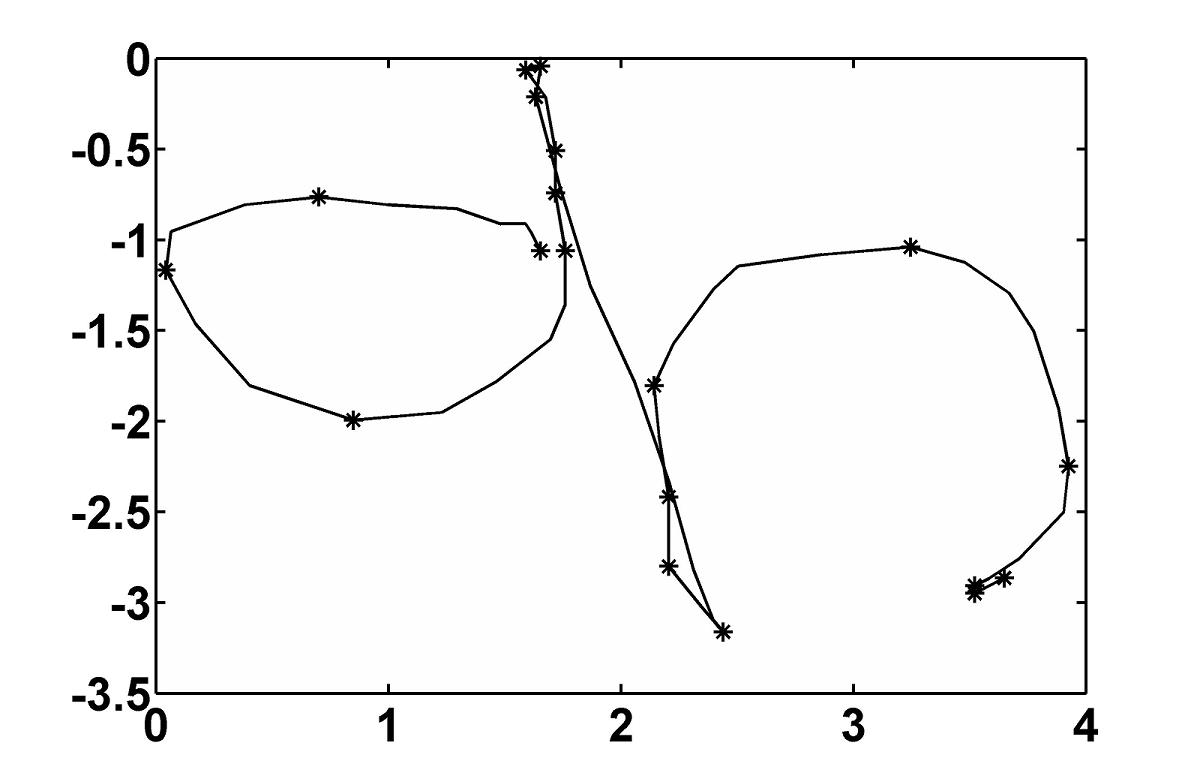}}
\caption{(a)-(b) Curvature points on on-line Devanagari characters \charaa\ and
\charka : Filtered using DWT }
\label{Fig:crt_pnt_dwt}
\end{figure}

\begin{figure}
\centering
\subfigure[]{
\includegraphics[width=0.45\textwidth]{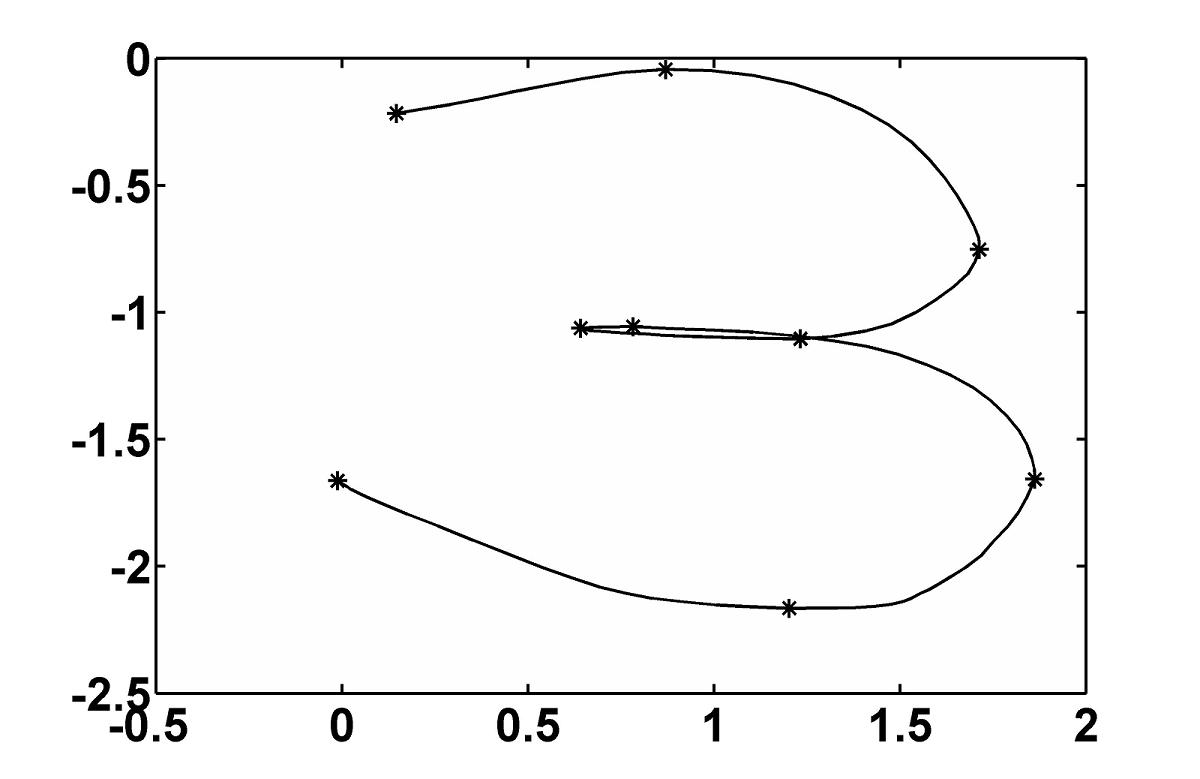}}
\subfigure[]{
\includegraphics[width=0.45\textwidth]{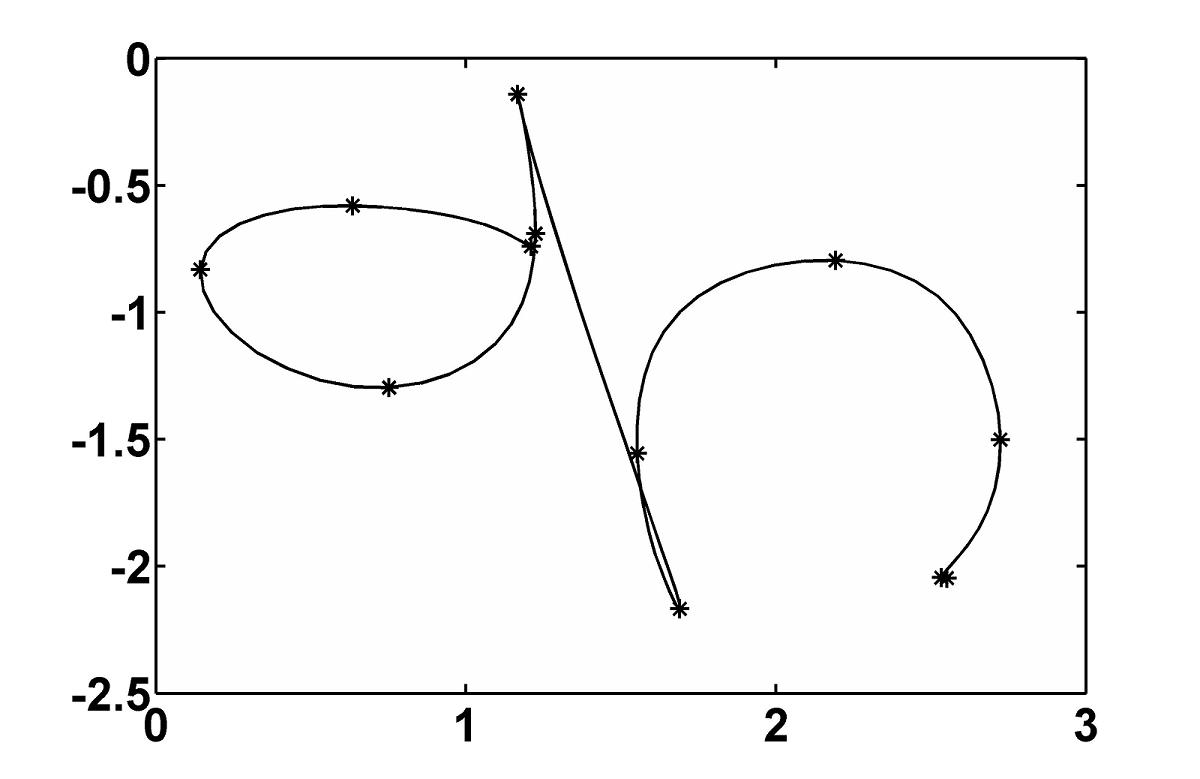}}
\caption{(a)-(b) Curvature points on on-line Devanagari characters \charaa\ and
\charka: Smoothed
using knotless spline method}
\label{Fig:crt_pnt_supsmu}
\end{figure}

We use $x'$ to compute the critical
point in $x$ sequence. The point $i$ is a critical point {\em iff}  $ x'_i - x'_{i+1}
\ne 0 $. Observe that $ x'_i - x'_{i+1}$ is the second difference. Similarly we calculate the critical points for the $y$ sequence. The final list
of critical points is the union of all the points marked as critical points in
both the $x$ and the $y$ sequence (see 
\Figure\ \ref{Fig:crt_pnt_org}, \ref{Fig:crt_pnt_dwt} and \ref{Fig:crt_pnt_supsmu}).
It must be noted that  the position and number of \cps\  computed for
different samples of the same strokes may vary significantly especially in the
presence of noise. \Figure\ \ref{Fig:crt_pnt_org} shows the \cps\ identified from the original
samples of on-line Devanagari characters \charaa\ and \charka.
 
\Figure\ \ref{Fig:crt_pnt_dwt} (a) and (b) shows the
\cps\ identified from the DWT filtered
on-line Devanagari characters \charaa\ and \charka\ respectively,
while \Figure\ \ref{Fig:crt_pnt_supsmu} (a) and (b) shows the \cps\
identified from the on-line Devanagari characters \charaa\ and \charka\
smoothed using the knotless spline method. 
It is clear from the \cps\ identified that the 
number and position of the \cps\ are more consistent and occur at the points
where there is  a change in curvature
when we apply the knotless spline based smoothing when compared to the \cp\
extraction on the raw data or the DWT. 

\begin{figure}
\centering
\includegraphics[width=0.90\textwidth]{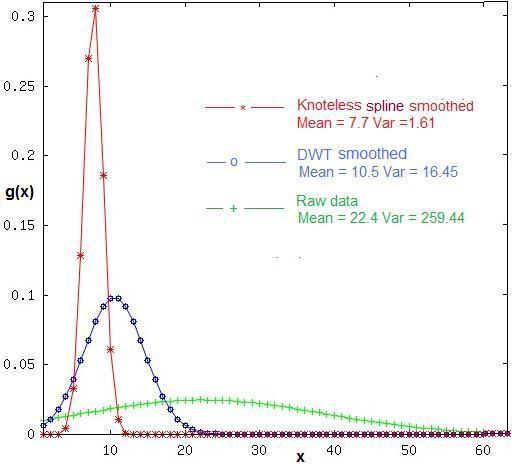}
\caption{Distribution curves for number of \cps\ obtained for (a) raw  (b) DWT smoothed  
(c) knotless spline smoothed Devanagari character \charaa.}
\label{fig_crt_pt}
\end{figure}

To understand the effect of smoothing on the extraction of \cps\ we conducted
an experiment on the \primitive\ data. The number of \cps\  for \mynumber\
samples of $10$ character primitives 
were computed and compared. Table \ref{tab_mv} shows the mean 
and variance of the number of \cps\ obtained for a sample $10$ primitives 
(a) without smoothing (raw data) (b) with DWT based smoothing (c) with knotless spline based smoothing. 
It is clear that the mean values obtained for knotless spline based smoothed data are very close 
to the expected values of the number of critical points, for respective primitives, compared to the mean values  
obtained for the raw data and the DWT smoothed data. The variance in the number of critical
points also very less in the case of knotless spline based smoothed data. Which
means that knotless smoothing helps in identifying close to the actual number
of \cps\ in a \primitive.
\begin{table}
\caption{Mean ($\mu$) and Variance ($\sigma^{2}$) of the Number of \cps}
\begin{center}
$
\begin{array}{||c|c|c|c|c|c|c||}
\hline
\mbox {Primitive} & 
\multicolumn{2}{|c|} {\mbox{Raw}}  & 
\multicolumn{2}{|c|} {\mbox{DWT}}   & 
\multicolumn{2}{|c||} {\mbox{Spline}} \\ \hline
& {\mu}&{\sigma^{2}} &{\mu}&{\sigma^{2}}&{\mu}&{\sigma^{2} }\\ \hline
 \mbox{\charaa\ } &22.4 &259.44 & 10.5& 16.45& 7.7&1.61\\
 i &19.9 &103.29 & 9.7 & 1.81 & 7.7&0.41\\
 e &13.9 &63.69  & 5.2 & 2.56 & 3.4&0.84\\
 k &29.3 &113.21 & 14.9& 7.49 & 11.9&1.49\\
 R &12   &23.8   & 5.6 & 0.44 & 4.4&0.24\\
 v &18.4 &64.64  & 9.2 & 3.96 & 7.6&1.44\\
 g &14.4 &53.44  & 6.8 & 1.76 & 5.5&0.45\\
 gh&19.6 &167.56 & 13.1& 4.29 & 9.8&2.36\\
 D &32.7 &88.44  & 7.1 & 6.89 & 5.6&3.24\\
 c&20.3&136.01  & 13  & 10.4 & 9.6&8.44\\ \hline
\end{array}
$
\end{center}
\label{tab_mv}
\end{table}
The distribution curves obtained for the primitive \charaa\ for (a) raw data 
($\mu = 22.4$ and $\sigma^{2} = 259.44$),
(b) DWT smoothed data ($\mu = 10.5$ and $\sigma^{2}=16.45$) and (c) knotless spline based smoothed data ($\mu = 7.7$ and $\sigma^{2}=1.61$) 
are shown in \Figure\ \ref{fig_crt_pt}. 
From the distribution plot it is clear that the knotless spline based smoothing can be effectively 
used for noise removal and hence improve the efficiency of the feature extraction process.

\subsection{Directional and Extended Directional Features}
\label{sec:df}
\begin{table}
\caption{Extended Directional Features}
\begin{center}
$
\begin{array}{|| c | c | c | c | c | c | c | c | c | c ||} \hline
\mbox{CP} & c_1 & c_2 & c_3 & c_4 & \cdots& \cdots & c_m &\cdots &  c_k \\ \hline
c_1&0&d_{12}&d_{13}&d_{14}&\cdots&\cdots&d_{1m}&\cdots&d_{1k} \\
c_2&-&0&d_{23}&d_{24}&\cdots&\cdots&d_{2m}&\cdots& d_{2k}\\
c_3&-&-&0&d_{34}&\cdots&\cdots&d_{3m}&\cdots& d_{3k}\\
c_4&-&-&-&0&&&&& d_{4k}\\
\vdots &\vdots&\vdots&\vdots&\vdots&\vdots&\vdots&\vdots&\vdots&\vdots \\
c_l&-&-&-&-&&&d_{lm}&& d_{lk}\\
\vdots &&&&&&&&& \\
c_k&-&-&-&-&-&-&-& -&0\\ \hline
\end{array} 
$
\end{center}
\label{fig:edf}
\end{table}

In the case of directional features (DF) we compute the
directions between only the adjacent \cps. 
%namely, $m = l+1$ in (\ref{eq:angle}). %(see Figure \ref{fig:df}). 
Suppose we have $k$ \cps, then we will obtain the direction 
between the consecutive \cps by first determining the angles
[$\theta_{12},$ $\theta_{23},$ $\theta_{34},$ $\cdots,$ $\theta_{k-1 k}]$ and then
[$d_{12},$ $d_{23},$ $d_{34},$ $\cdots$ $d_{k-1 k}$ ] using Algorithm \ref{algo:cp2dir}. 

The extended directional feature (EDF) set is computed by computing all 
direction between one \cp\ and all other \cps\ following it. % as shown in Figure \ref{fig:edf_angles}.
These directions are indicated as the upper diagonal elements in Table
\ref{fig:edf}. Where $d_{lm}$ 
corresponding to the angle $\theta_{lm}$ (computed using the Algorithm
\ref{algo:cp2dir}) and is the direction between the \cp\  $c_l$ and $c_m$.
% \item 
Given, $k$ \cps, we get an extended directional feature (EDF) vector of size
\begin{equation} \frac{k(k-1)}{2} \end{equation} while we get a directional feature (DF) vector of size $k$.

\subsection{Fuzzy Directional Feature Extraction}
\label{sec:fdf}

We propose a simple yet effective feature set based on fuzzy directional feature set\footnote{Note that \cite{mukherji} 
talks of fuzzy feature set
for Devanagari script albeit for off-line handwritten character recognition}.
Let $k$ be the number of \cps\ 
(denoted by $c_1, c_2, \cdots c_k$) extracted from a stroke of length $n$; usually $k << n$.
The $k$ critical points form the basis for extraction of the directional features and the FDF.
We first compute the angle between two critical points, say $c_l$ and $c_{l+1}$, as
\begin{equation}
\theta_{l} = \tan^{-1}\left (\frac{y_l - y_{l+1}}{x_l - x_{l+1}} \right )
\label{eq:angle}
\end{equation}
where $(x_l ,y_l)$ and $(x_{l+1},y_{l+1})$ are the coordinates corresponding
to the \cp\ $c_l$ and $c_{l+1}$ respectively.

\begin{figure}
\centering
\includegraphics[width=0.9\textwidth]{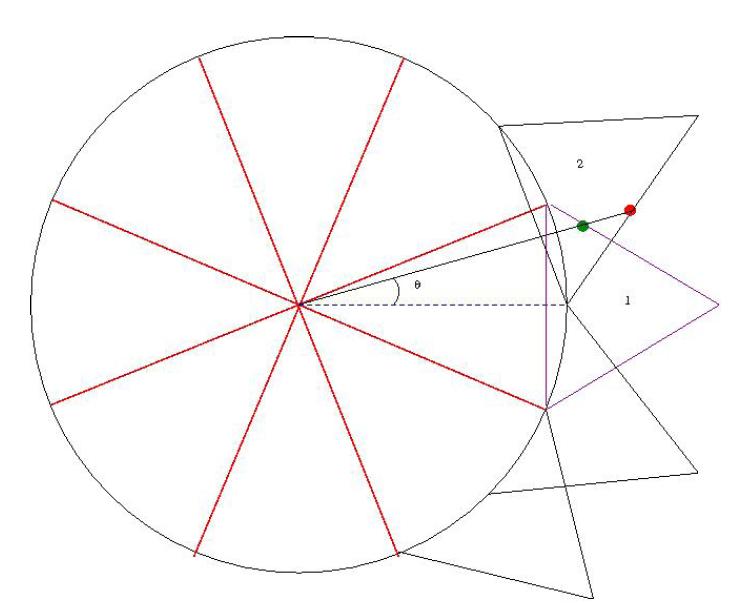}
\caption{$\theta$ contributing to two directions $(1, 2)$ with fuzzy membership values (green and red dot)}
\label{fig_angles}
\end{figure}

Note that we divide $2\pi$ into \eight\ directions with overlap. 
Every $\theta_{l}$ (for example the angle $\theta$ that 
the blue dotted line makes with the horizontal axis in \Figure\ \ref{fig_angles}) has two directions (say $ d^1_{l} = 1, d^2_{l} = 2$, note that the
line making an angle $\theta$ with the dotted line 
in \Figure\ \ref{fig_angles} lies in both the triangles represented by direction $1$ and direction $2$) associated
with it having $m^1_{l}, m^2_{l}$ membership values respectively (represented
by the green and the red dot respectively in \Figure\ \ref{fig_angles}. 
Also note \begin{enumerate}
\item $m^1_{l}+m^2_{l} =1$ and \item $d^1_{l}, d^2_{l}$ are adjacent
directions,
for example
if $d^1_{l} = 5$ then $d^2_{l}$ could be either $4$ or $6$.
				\end{enumerate} 
\begin{algorithm}
\begin{algorithmic}
\STATE{int deg2dir(double $\theta$)}
\STATE{int dir = -1;}
\IF{($\theta > -\pi/8$ \& $\theta < \pi/8$)}
	\STATE{dir = 1;}
\ENDIF
\IF{($deg >= \pi/8$ \&  $\theta < 3 \pi/8$) }
	\STATE{dir = 2;}
\ENDIF
\IF{($\theta >= 3 pi/8$ \&  $\theta < 5 pi/8$) }
	\STATE{dir = 3;}
\ENDIF
\IF{($\theta >= 5 \pi/8$ \& $\theta < 7 \pi/8$) }
	\STATE{dir = 4;}
\ENDIF
\IF{(($\theta >= 7 \pi/8$ \& $\theta < 9 \pi/8$) $\|$ ($\theta >= -9 \pi/8$ \& $\theta
< -7 \pi/8$))}
	\STATE{dir = 5;}
\ENDIF
\IF {($\theta >= -7 \pi/8$ \& $\theta < -5 \pi/8$) }
	\STATE{dir = 6;}
\ENDIF
\IF {($\theta >= -5 \pi/8$ \& $\theta < -3 \pi/8$) }
	\STATE{dir = 7;}
\ENDIF
\IF{($\theta > -3 \pi/8$ \& $\theta < -\pi/8$) }
	\STATE{dir = 8;}
\ENDIF
\STATE{return(dir);}
\end{algorithmic}
\caption{Angle between two \cp\ conversion into \eight\ direction}
\label{algo:cp2dir}
\end{algorithm}

\begin{algorithm}
\begin{algorithmic}
\STATE {fuzzy-membership($\theta_{c} ,\theta$);}
\STATE {$m = 1.0 - \frac{(|(\theta_{c} - \theta)|)}{\pi/4};$}
\STATE {return(m);}
\end{algorithmic}
\caption{Triangular Fuzzy Membership Function}
\label{algo:fuzzy_tmf}
\end{algorithm}

We use $\theta$ in Algorithm \ref{algo:cp2dir} assisted by triangular membership 
function described in 
Algorithm \ref{algo:fuzzy_tmf} for computing the FDF set (refer Table \ref{fig_fdf}). 
Here $\theta_{1,k-1}$ is the 
angle between two consecutive \cps\ (where $k$ is the total number of \cps) in a handwritten primitive and 
$d_{1,8}$ is the respective direction. The fuzzy membership values assigned to each direction are represented as $m^{1,2}_{1,k-1}$ and
the corresponding feature vector values as $f_1,\ldots,$ and $f_{8}$. It should be noted that the sum of the membership functions of
a particular row in Table \ref{fig_fdf} ) is always $1$. We calculate the FDF by averaging across the columns, so as to form a vector 
of dimension eight. The mean is calculated as follows; for each direction (1 to
8), collect all the membership values and divide by 
the number of occurrences of the membership values in that direction. For example, in Table \ref{fig_fdf}, the mean for direction $1$
is calculated as \begin{equation} f_1 =
\frac{(m^{1}_{2}+m^{1}_{3})}{2}\end{equation} 
In all our experiments we have used these mean values to construct
$8$ directional FDF to represent a stroke. %(refer \ref{eq_fv}).

Clearly, the fuzzy aspect comes into picture due to the membership function which associates
the angle between two \cps\ into two directions with different membership values. In the commonly used 
Directional Features (see Section \ref{sec:df}) only one direction is associated with each $\theta$ (the
angle between two consecutive \cps).

\begin{table}
\caption{Computation of Fuzzy Directional Features}
\begin{center}
$
\begin{array}{c|c|c|c|c|c|c|c|c}
\mbox{$ \theta \downarrow)$ $ d \rightarrow )$} & 1 & 2 & 3 & 4 & 5&  6&  7&  8 \\ \hline
 &&&&&&&& \\
\theta_{1} &&&m^{1}_{1}&m^{2}_{1}&&&&\\
\theta_{2} &m^{1}_{2}&m^{2}_{2}&&&&&&\\
\theta_{3} &m^{2}_{3}&&&&&&&m^{1}_{3}\\
\vdots &&&&&&&& \\
%p&&&&&m_{p1}&m_{p2}&& \\
\theta_{l} &&&&&&&m^{2}_{l}&m^{1}_{l}\\ 
\vdots &&&&&&&& \\
\theta_{k-1} &&&&&m^{2}_{k-1}&m^{1}_{k-1}&&\\ &&&&&&&&\\ \hline  &&&&&&&& \\
F & f_{1}=&f_{2}= &f_{3}=&f_{4}=&f_{5}=&f_{6}=&f_{7}=&f_{8}= \\
& \frac{(m^{1}_{2}+m^{1}_{3})}{2} &m^{2}_{2} & m^{1}_{1} & m^{2}_{1}&m^{2}_{k-1}& m^{1}_{k-1}&m^{2}_{l}
&\frac{(m^{1}_{3}+m^{1}_{l})}{2}
\end{array}
$

\end{center}
\label{fig_fdf}
\end{table}

%Recognition experiments and results

\section{Experimental Results}
\label{sec:experiments}
\newcommand{\Done}{{\em Dataset 1}}
\newcommand{\Dtwo}{{\em Dataset 2}}

Experiments were carried out for two different datasets, namely, (a) the
isolated primitive dataset (\Done) where the identified $69$ primitives were written by
\mynumber\ users and  (b) primitives extracted (\Dtwo) from the 
paragraph data (continuous text). Note that in \Done\ people were asked to just
write the strokes and hence was not natural while in \Dtwo\ the writers were
asked to write a paragraph using e-pen. 
We used knotless spline noise removal in all our experiments because this
performed the best as seen in earlier sections. 
Critical point were extracted from the smoothened \primitive\ and three different set of features 
were extracted for each of the dataset, namely,
Directional Features (DF), Extended Directional Features (EDF) and Fuzzy
Directional Features (FDF) to build the reference models for the $69$
\primitives. 
In the testing (recognition) phase we used three different types of
classifiers, namely (a) Second order statistics based nearest neighborhood classifier 
(b) Dynamic Time Warping (DTW) based
classifier and 
(c) Support Vector Machine (SVM) based classifier. 
The recognition results are then compared across the two dataset, three feature sets and
three classifiers.

\subsection{Recognition of Isolated primitive (\Done)}
\label{recog_prim}

The recognition experiments are conducted as follows. 
We collected $10$ isolated samples of each $69$ Devanagari primitive 
strokes from \mynumber\ different writers using Mobile e-Note Taker. 
We divide the collected data into two non-overlapping sets, 
namely the train dataset and test dataset. 
The training and test set were in the ratio of ${80:20}$ and {\em five-fold cross validation}
procedure is performed and the recognition accuracies computed.

During training, we calculate the three feature set (DF, EDF and FDF) for all the training samples 
corresponding to the same \primitive. These features are extracted after knotless spline  
smoothing. The mean and variance feature vector is computed for each
\primitive\ from the feature vectors of that \primitive\ from
the training dataset.

For testing purpose, we took a stroke $t$ from the test dataset and extracted
the three features after noise removal. Then the distance
between the test feature vector and all the $69$ reference \primitives\ is
computed using second order statistics based classifier, Discrete Time Warping (DTW) 
and the Support Vector Machine (SVM) based classifier as described below.

\subsubsection{\Done\ using second order statistics}
\label{recog_second_stat}
In case of FDF for all stroke corresponding to the same \primitive\ as
described in Section \ref{sec:fdf}. Then we computed the mean $(\mu)$ and
co-variance ($\Sigma$) for each of the $69$ \primitive\ to model the primitive.
If there are $\beta$ strokes corresponding to a \primitive, then we have $\beta$ $F$s, 
say, $F_{1}, F_{2}, \ldots, F_{\beta}$ 
then each primitive is modeled as
\begin{equation}
\mu = \frac{1}{\beta}\sum_{i=1}^{\beta} F_{i}
\label{eq_fv2}
\end{equation}
\begin{equation}
\Sigma = \frac{1}{(\beta - 1)} \sum_{i=1}^{\beta} (F_i - \mu) {(F_i - \mu)}^T
\label{eq_covar}
\end{equation}

Here a primitive is represented by a mean vector $(\mu)$ of size $1 \times 8$ and co-variance matrix of size $8\times8$. The class reference models  
are represented as ${(\mu_i,\sum_i)}^{69}_{i=1}$.
For testing purpose, we took a stroke $(t)$ from the test data set and extracted FDF as described in Section \ref{sec:fdf} and compared it with
$69$ reference models. The likelihood of the test primitive $t$ with reference primitive $k$ (where $k = 1, 2, \cdots, 69$) 
is computed as
\begin{equation}
P(t/k) = \left [ \frac{1} {{(2\pi)}^{\frac{8}{2}} \sqrt {|\sum_k|}} \right ]  \exp^{-\frac{1}{2} (t-\mu_{k}) \Sigma^{-1}_k {(t-\mu_{k})}^{T}}
\label{eq_stat}
\end{equation}
The $k$ for which (\ref{eq_stat}) is maximum is identified as the recognized \primitive.

\begin{table}
\begin{center}
\caption{Recognition results for different $\alpha$}
\label{tab_res3}
\begin{tabular}{||l|c|c|c|c|c|c|c|c||} \hline
Feature Set&Data&$\alpha=1$&$\alpha=2$&$\alpha=5$\\ \hline
FDF&Train&80.25\% &85.46\%&96.89\%\\ %\cline{1-8}
&Test&70.29\%&74.86\%&82.75\%\\ \hline
\end{tabular}
\end{center}
\end{table}

Table \ref{tab_res3} shows the experimental results obtained for both the 
train and the test data for FDF using second order statistics based classifier.
Note that the values in Table \ref{tab_res3} are computed as follows. For $N = \alpha$, the test stroke $t$ is recognized as the primitive 
$l$ if $l$ occurs at least at the $\alpha^{th}$ position from the best match (this is generally called the N-best in literature).
It should be noted that the recognition accuracies are for writer independent unconstrained strokes.
As expected the recognition accuracies are not very high (very similar to the phoneme recognition in
speech literature) for $\alpha =1$ but improves with increasing $\alpha$. It is also noted that similar experiments 
with Directional Features (DF) shows $\pm 10\%$ lower recognition efficiency. 
The recognition experiments are also performed using DTW, SVM classifiers which are explained in following sessions. 

\subsubsection{\Done\ based on DTW}
\label{recog_dtw}

Being an elastic matching technique, Dynamic Time Warping (DTW)\footnote{edit distance in Computer Science literature}  
allows to compare two sequences of different lengths. 
This is especially useful to compare patterns in
which rate of progression varies non-linearly, which makes similarity measures such as Euclidean distance and cross-correlation
unusable \cite{joshi}. 
Since different strokes corresponding to the same \primitive\ had
different feature vector length in the case of Directional Features (DF) and  
Extended Directional Features (EDF)
we used DTW algorithm\footnote{We do not discuss this
algorithm, in detail since it is well documented in literature} to compute the
distance between the test \primitive and the reference \primitives.
We calculated the DTW distance of the test \primitive\ from all the reference
\primitives. between all strokes corresponding to the same 
\primitive\ in the training set with the test stroke. The 
stroke with minimum average stroke is identified as the recognized primitive. The recognition experiments are conducted based on
DF and EDF ( which are unequal length feature vectors) and recognition accuracies are tabulated \ref{tab_res1}.

\subsubsection{\Done\ based on Support Vector Machine}
\label{recog_svm}

The principle of an SVM is to map the input data
onto a higher dimensional feature space non linearly related
to the input space and determine a separating hyper-plane
with maximum margin between the two classes in
the feature space \cite{cjc}.
This results in a nonlinear boundary
in the input space. The optimal separating hyper-plane can
be determined without any computations in the higher dimensional
feature space by using kernel functions in the
input space. We used a multi-class SVM classifier with a Radial Basis Function (Gaussian) Kernel function \cite{Bahlmann} 
for our experimentation.   
Since data should have a constant length representation in the SVM-based
approach, FDF (a fixed length feature vector) have been used in the recognition experiments. 
The recognition results are tabulated. 
The recognition results for DF, EDF and FDF are compared. It can be 
observed that the FDFs give  better recognition accuracies  and
SVMs can be used effectively for \primitive\ recognition (see Table
\ref{tab_res1}).

\begin{table}
\begin{center}
\caption{Recognition results for \Done (in \%)}
\label{tab_res1}
\begin{tabular}{||c|c|c|c|c|c|c||} \hline
Classifier&\multicolumn{2}{|c|} {DF}& \multicolumn{2}{|c|} {EDF}&
\multicolumn{2}{|c|} {FDF} \\ \hline
&Train&Test&Train&Test&Train&Test\\ \hline
Second Order &65.22 &60.15 &69.02 &66.66 &80.25 &70.29 \\  \hline
DTW&61.41 &58.67 &72.28 &67.39 &- & -\\ \hline
SVM&69.20 &63.04 &77.72 &68.84 &86.95 &73.91 \\ \hline
\end{tabular}
\end{center}
\end{table}

\subsection{Recognition of \primitives\ in Paragraph (\Dtwo)}
\label{recog_para1}

We used $80$ \% of \mynumber\ user paragraph data for training and the rest for the purpose of testing
the performance of the ED feature set and the DF set. 
We initially extracted the \primitives\ from the paragraph dataset and hand
tagged each stroke as belonging to one of the $69$ \primitives\
(see \Figure\ \ref{fig_primitives}). 
All the strokes corresponding to the given \primitive\ in the
training data was collected and clustered together. 
For each stroke we extracted the DF, EDF and FDF set as described in Section
\ref{sec:fdf} and \ref{sec:df}. 
All strokes corresponding to the same \primitive\ which were within a distance of $\tau$ were clustered together 
and only one representative stroke from the cluster was retained as the cluster representative. 

For testing purpose, we took a stroke ($t$) from the test data, we first
extracted EDF\footnote{for experiments with DF, we extracted the corresponding
directional features} and compared it with
the EDF of the all reference strokes using DTW 
algorithm. We then took the average distance of $t$ from all
the references of a \primitive. We arranged these average distances ($69$ in
number) in the increasing order of magnitude. The primitive with the least
average distance from the test stroke $t$ is declared as being recognition of
stroke $t$.

%Table \ref{tab_data} shows the number of strokes used as the train and test 
%data set corresponding to the $20$ selected primitives.  Note that these train
%and test data\footnote{hand tagged} come
%from the paragraph data written by different people.
%\begin{table}
%\caption{Recognition accuracies for isolated stroke dataset.}
%\label{tab_data}
%\begin{center}
%\begin{tabular}{||l|c|c|c|c|c||} \hline
%%&&  Training Set&               & Recognition Results&\\
%Primi-&Training& Test& DF&EDF&FDF\\ 
%tives&Samples &Samples & & &\\ \hline
%R & 82 &71 &66 & 46 &56\\
%l & 69 &18 &9 &7 & 10\\
%K & 69 &57 &55 & 51&52\\
%N & 63&67 & 33& 36&35\\
%V & 50&36 & 19& 23 &27\\
%p & 58&36 &31 &32 &32\\
%dd & 54&50 &19 & 34&36\\
%m & 39&50 & 15&36 &25\\
%tt & 28&23 & 8&11&14 \\
%y & 40&36& 16& 22 &24\\
%a & 33&28& 25&25& 17\\
%h &34&16 &9 &9 &13\\
%T & 26&25 &12 &15&18\\
%g &24&25& 13& 10& 13\\
%D & 22&21 & 14&13&15\\
%e & 15&16 &3 &7&8\\
%J & 17&18 & 13&13&13\\
%ii & 12&10 &4 &4& 5\\
%ch & 10&11 &10 &9&9\\
%C & 10&11 &4 &7&7\\ \hline
%Total &745& 625&378 &410 &429 \\ \hline
%\end{tabular}
%\end{center}
%\end{table}
%
%The overall stroke level recognition
%accuracies for  DF, EDF  and FDF obtained using second order statistics based classifier are 
%%(refer Table \ref{tab_data})
%$60.48$\% ,  $65.6$\% and $68.64$\% respectively. 
%It should be noted that there is an improvement in the recognition accuracies
%due to use of FDF.
The recognition results obtained for different features sets using different 
classification techniques are tabulated in Table \ref{tab_res2}.
\begin{table}
\begin{center}
\caption{Recognition results for \Dtwo}
\label{tab_res2}
\begin{tabular}{||c|c|c|c||} \hline
Classifier&DF&EDF&FDF \\ \hline
Second Order &63.68 \% &65.60 \% &68.64 \% \\ \hline
DTW&57.92 \%&66.88 \%& - \\ \hline
SVM&61.12 \%&67.36 \%&71.73 \% \\ \hline
\end{tabular}
\end{center}
\end{table}
The results obtained for recognition of Devanagari primitive strokes 
show that reliable classification is possible using
SVMs and FDF can be easily extended
to other Indian scripts as well. The results also indicate the scope for further
improvement. Future work is directed towards extending the stroke level recognition to 
character level recognition and further to world level recognition based on the 
proposed framework in Section \ref{sys_arch}.

\section{Conclusion}
\label{sec:conclude}

In this paper we have presented an on-line handwriting script recognition
framework for Devanagari which can be extended to other Indian language 
scripts.  The framework is motivated by speech recognition literature and in
our opinion has not been reported in literature.
%An end to end OHCR system architecture is also proposed 
%for Indian languages. A quick review on the OHCR work done on Indian 
%language is also presented. The isolated character data and the 
%paragraph data are collected using e-Pen and prepared a Devanagari 
%stroke database for experimentation.
Handwritten script is inherently noisy and we experimented with several noise
removal techniques and identified that the noise removal is a necessary
pre-processing step in OHCR, we further showed that
knotless spline based denoising work better than the wavelet based denoising. 
%The analysis on the 
%number of \cps on the smoothed data indicates that the number of 
%critical points and the positions on the stroke curves are very close to 
%the expected values.
%on-line handwriting data. 
%The 
%proposed method is tested and evaluated on Devanagari on-line 
%handwritten strokes and found that it has the ability to remove noise 
%much better than the DWT based denoising method. We also compared the 
%recognition results on raw data (without pre-processing), DWT based 
%denoised data set and the data smoothed using the proposed method. It is 
%evident, from the recognition results, that the performance of the 
%recognition algorithm significantly improves when data is smoothed using 
%proposed technique.
We introduced a feature set
% extraction stage we introduced a new on-line script 
%feature set, 
called the Fuzzy Directional Features (FDF) which is able to incorporate
directional variance in the handwritten \primitives. 
%As a 
%pre-processing to the feature extraction we have identified critical 
%points (high curvature) points on each stroke. 

Experimental results show that the performance of the FDF  
for the two datasets is better that for other feature sets (DF and EDF).
The recognition accuracies obtained for \Done\ (isolated \primitive) 
was 73.91\% (using FDF and SVM classifier) and for \Dtwo\ (\primitives\ extracted 
from unconstrained continuous script data) was 71.73 \%.
% using FDF and SVM 
%classifier). 
Experiments show that   
FDF performs much better than the commonly used directional 
features.
%level recognition shows that reliable classification is possible using 
%SVMs and can be easily extended to other Indian scripts as well.

Character based OHCR reported in literature fail because of the
increased number of characters in the Indian language script and the presence
of a large number of complex compound characters. The advantage of the 
proposed framework based
on stroke (\primitive) recognition rather than character recognition is its
ability to
handle complex characters. It is  because the number of \primitives\ to be recognized do not increase
with increased character set.  
%Further, the framework allows 
%One of the problems in Indian script is the recognition of complex compound 
%characters. 
% Much research is needed on recognition of complex characters and 
%complete words in the different Indian scripts and the proposed OHCR 
%framework will lead to fulfill the goal. As continual work we plan to 
%use (a) the spatio-temporal information to enhance character(constitute 
%of multiple strokes) recognition, (b) HMMs and lexicon based information 
%for word level recognition. Since the stroke order, direction and number 
%variations play a crucial role in the recognition of Indian scripts, 
%offline features will also definitely help for improving recognition 
%accuracy when used in combination with on-line features and approaches.

%It is also noted that there is an urgent need for standardized datasets 
%to be created and made available for the research community for Indian 
%languages, it will also help in meaningful comparison of different 
%published approaches. This we believe will lead to better writer 
%independent on-line handwriting recognition system for Indian languages.

\section{Acknowledgements}
The authors would like to thank Dr Vinod Pandey for sharing the 
implementation of the spline smoothing algorithm reported in this paper. 
% If you'd like to thank anyone, place your comments here
% and remove the percent signs.
%\end{acknowledgements}


\bibliographystyle{plain}      % basic style, author-year citations
\bibliography{ohcr}   % name your BibTeX data base

  %\begin{thebibliography}{}
% etc
 %\end{thebibliography}

\end{document}